\documentclass[11pt]{article}

\usepackage[preprint]{acl}

\usepackage{times}
\usepackage{latexsym}

\usepackage[T1]{fontenc}

\usepackage[utf8]{inputenc}

\usepackage{microtype}

\usepackage{inconsolata}

\usepackage{graphicx}

\usepackage{hyperref}
\usepackage{url}
\usepackage{algorithm}
\usepackage{algorithmic}
\usepackage{subfiles}
\usepackage{booktabs}       
\usepackage{amsmath,mathtools}
\usepackage{amssymb}
\usepackage{amsfonts}

\usepackage{multirow}
\usepackage{multicol}
\usepackage{xcolor}
\usepackage{adjustbox}
\usepackage{amsthm}
\theoremstyle{definition}
\newtheorem{definition}{Definition}

\theoremstyle{remark}

\usepackage{pifont}
\usepackage{caption}
\usepackage{subcaption}
\usepackage{wrapfig}
\usepackage{simplebnf}
\usepackage{tcolorbox}
\usepackage{tikz}
\usepackage{tikz-qtree}
\usepackage{dashrule}
\tcbuselibrary{listings,breakable}
\tcbset{listing engine=listings,colframe=black,colback=white,size=small}
\usepackage{enumitem}

\usepackage{booktabs,multirow,tabularx,makecell,array}
\newcolumntype{Y}{>{\raggedright\arraybackslash}X} 

\usepackage[table]{xcolor}

\definecolor{lightgreen}{RGB}{220,245,220}
\definecolor{lightred}{RGB}{255,220,220}

\newcommand{\cmark}{\cellcolor{lightgreen}\checkmark}
\newcommand{\xmark}{\cellcolor{lightred}\texttimes}

\title{ReviewScore: Misinformed Peer Review Detection \\with Large Language Models}

\author{
 \textbf{Hyun Ryu\textsuperscript{1,2}},
 \textbf{Doohyuk Jang\textsuperscript{1}},
 \textbf{Hyemin S. Lee\textsuperscript{3}},
 \textbf{Joonhyun Jeong\textsuperscript{1}},
 \textbf{Gyeongman Kim\textsuperscript{4}},
\\
 \textbf{Donghyeon Cho\textsuperscript{1}},
 \textbf{Gyouk Chu\textsuperscript{1}},
 \textbf{Minyeong Hwang\textsuperscript{1}},
 \textbf{Hyeongwon Jang\textsuperscript{1}},
 \textbf{Changhun Kim\textsuperscript{5}},
\\
 \textbf{Haechan Kim\textsuperscript{1}},
 \textbf{Jina Kim\textsuperscript{1}},
 \textbf{Joowon Kim\textsuperscript{1}},
 \textbf{Yoonjeon Kim\textsuperscript{1}},
 \textbf{Kwanhyung Lee\textsuperscript{1,5}},
\\
 \textbf{Chanjae Park\textsuperscript{1}},
 \textbf{Heecheol Yun\textsuperscript{1}},
 \textbf{Gregor Betz\textsuperscript{6}},
 \textbf{Eunho Yang\textsuperscript{1,5}}
\\
\\
 \textsuperscript{1}KAIST,
 \textsuperscript{2}Carnegie Mellon University,
 \textsuperscript{3}MIT,
 \textsuperscript{4}KRAFTON,
 \textsuperscript{5}AITRICS,
 \textsuperscript{6}KIT
\\
 \texttt{\{ryuhyun1905,eunhoy\}@kaist.ac.kr, hyunr@cmu.edu}
}

\begin{document}
\maketitle
\begin{abstract}
Peer review serves as a backbone of academic research, but in most AI conferences, the review quality is degrading as the number of submissions explodes. To reliably detect low-quality reviews, we define \textit{misinformed} review points as either ``weaknesses'' in a review that contain incorrect premises, or ``questions'' in a review that can be already answered by the paper. We verify that 15.2\% of weaknesses and 26.4\% of questions are \textit{misinformed} and introduce \textsc{ReviewScore} indicating if a review point is \textit{misinformed}. To evaluate the factuality of each premise of weaknesses, we propose an automated engine that reconstructs every \textit{explicit} and \textit{implicit premise} from a weakness. We build a human expert-annotated \textsc{ReviewScore} dataset to check the ability of LLMs to automate \textsc{ReviewScore} evaluation. Then, we measure human-model agreements on \textsc{ReviewScore} using eight current state-of-the-art LLMs. The models show F1 scores of 0.4--0.5 and kappa scores of 0.3--0.4, indicating moderate agreement but also suggesting that fully automating the evaluation remains challenging. A thorough disagreement analysis reveals that most errors are due to models' incorrect reasoning. We also prove that evaluating premise-level factuality shows significantly higher agreements than evaluating weakness-level factuality.\footnote{We will make the source code and dataset publicly available.}
\end{abstract}

\section{Introduction}
\label{sec:intro}
Peer review is an evaluation of academic work by experts to assess its originality, significance, and validity before publication~\citep{peer1, peer-web1}.
In AI conferences, as the number of submissions is exponentially increasing, required number of reviewers is also exploding.
As a result, review quality is degraded, which undermines the integrity and reliability of a peer-review system~\citep{Stelmakh2021NoviceReviewer, CortesLawrence2021Inconsistency, Shah2022CACM, position}.
Due to the importance of the issue, previous works propose criteria for evaluating review quality. However, we observe a trade-off between applicability to reviews and specificity of rubrics. \citet{goldberg} and \citet{review-critique} proposes criteria that could apply to nearly every review but those are quite vague and subjective. \citet{guo}, \citet{sadallah}, \citet{purkayastha}, and \citet{claimcheck} introduce specific and objective criteria but those target narrow scope of reviews.

\begin{figure}[t]
  \centering
  \includegraphics[width=\columnwidth]{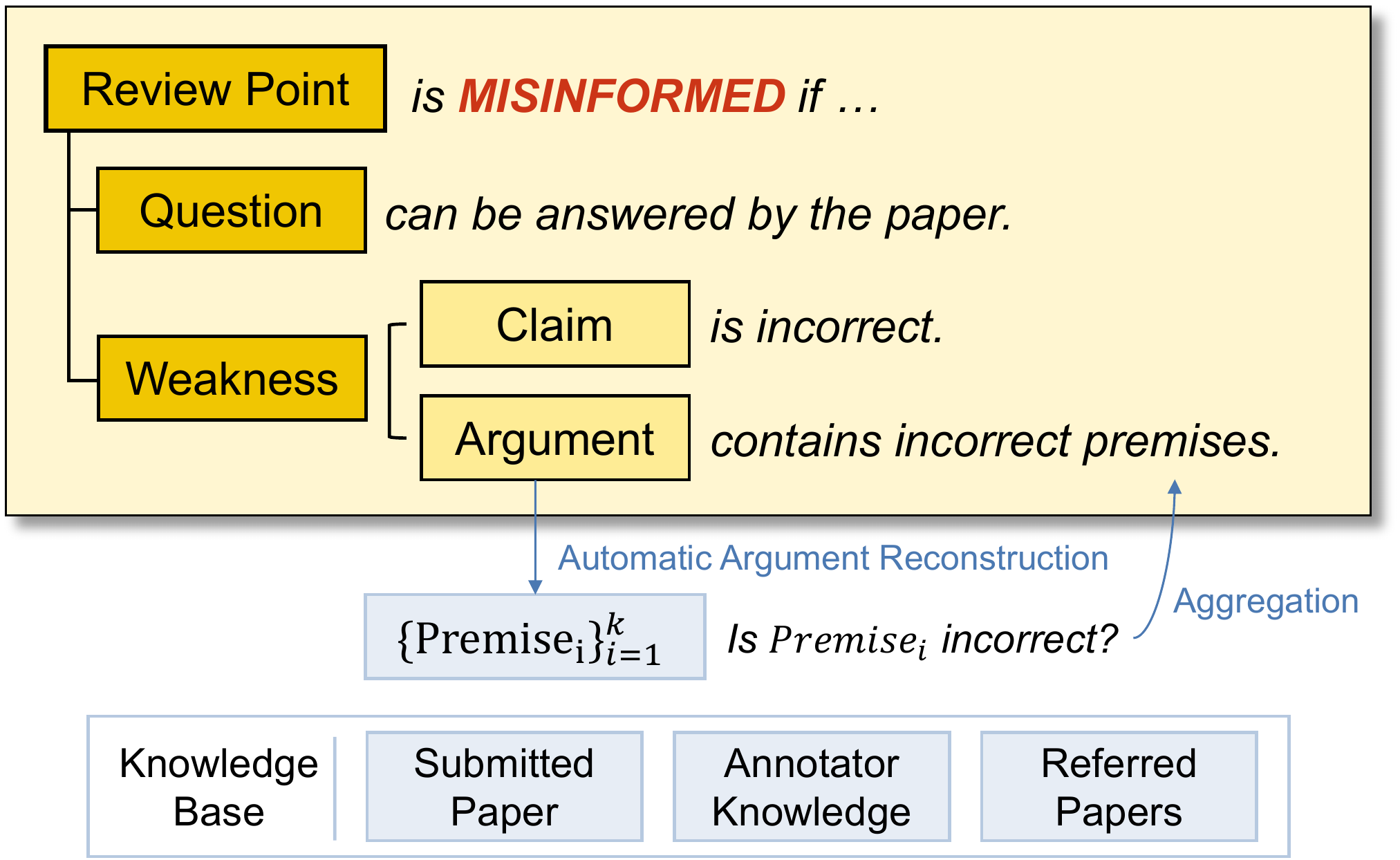}
  \caption{
  Overview of \textsc{ReviewScore}. Each review point in a review is categorized into question and weakness. We further categorize weakness into claim and argument by the presence of supporting reasons.
  Based on an appropriate knowledge base, if a question is answerable by the paper, a claim is factually incorrect, or an argument contains factually incorrect premises, then the review point is \textit{misinformed}.
  For arguments, to extract all \textit{explicit} and \textit{implicit premises}, we also introduce an automatic argument reconstruction engine.
  }
  \label{fig:main}
  \vspace{-5mm}
\end{figure}

To resolve this issue, we introduce two specific yet applicable criteria of a review quality: \textit{unanswerability of questions} and \textit{factuality of weaknesses}.
To select the criteria, we recruited a group of human experts and let them independently analyze a small subset of ICLR reviews\footnote{A group of three graduate students studying AI analyze reviews of ICLR submissions in OpenReview. Detailed process of this group work is described in Appendix \ref{sec:criteria-selection}.}.
Specifically, each of them decomposed a review into \textit{review points}, which are formally defined as follows.
\begin{definition}[Review Point]
A review point is a single, self-contained unit of evaluation or inquiry in a review--either a weakness or a question--that stands on its own semantically.
\end{definition}
Each human evaluated quality of review points, and they discussed trustworthy criteria to detect low-quality.
Based on the discussion, we formally define a \textit{misinformed} review point as follows.
\begin{definition}[Misinformed Review Point]
A review point is \emph{misinformed} if and only if
\begin{itemize}[leftmargin=7mm]
\vspace{-2mm}
\item a question stated in a review can be already answered by the paper, or
\vspace{-2mm}
\item a weakness stated in a review is incorrect or contains incorrect premises regarding the paper.
\end{itemize}
\end{definition}

In this work, the human annotation shows that 26.4\% of questions and 15.2\% of weaknesses are \textit{misinformed}, which means that the current AI conference reviews contain considerable amount of \textit{misinformed} review points.\footnote{A group of 15 graduate students annotate reviews of 40 works submitted to ICLR 2021-2023. Detailed process of this human annotation is described in Section \ref{subsec:dataset}.}
We also note that these criteria aligns with reviewer guidelines of major AI conferences. For instance, ACL 2023 Peer Review Policies indicate that ``before writing a negative review, check whether your questions are already answered.''~\citep{ACL2023PeerReviewPolicies}, and NeurIPS 2025 reviewer guidelines indicate that ``reviewers should minimize the chance of misunderstandings during the reviewing process''~\citep{NeurIPS2025ReviewerGuidelines}.

\begin{table*}[t]
\vspace{-5mm}
\centering
\footnotesize
\begin{tabular}{lccccc}
\toprule
 & Target Unit & Groundedness & Factuality & Premise Factuality & (Un)Answerability \\
\midrule
\citet{shin2025mind}& Full Review  & \xmark & \xmark & \xmark & \xmark \\
\citet{purkayastha} & Weakness       & \xmark & \xmark & \xmark & \xmark \\
\citet{guo}         & Full Review  & \cmark & \xmark & \xmark & \xmark \\
\citet{sadallah}    & Weakness       & \cmark & \xmark & \xmark & \xmark \\
\citet{claimcheck}  & Weakness       & \cmark & \xmark & \xmark & \xmark \\
\citet{review-critique} & Weakness Sentence & \cmark & \cmark & \xmark & \xmark \\
\midrule
\textsc{ReviewScore} (Ours) & Review Point  & \cmark & \cmark & \cmark & \cmark \\
\bottomrule
\end{tabular}
\caption{Comparison of automatic review evaluation methods in five criteria. Full Review contains a set of strengths and weaknesses.}
\label{tab:eval_dimensions}
\vspace{-5mm}
\end{table*}

Based on these observations, we define \textsc{ReviewScore} that indicates if a review point is \textit{misinformed} (Section \ref{subsec:reviewscore-def}).
First, we define \textsc{Base ReviewScore} by directly applying the definition of \textit{misinformed} review points in a 5-point scale.
However, we discovered human-annotated factuality often diverges. This is because a weakness often contains both correct and incorrect \textit{explicit premises} or incorrect \textit{implicit premises}, which hinders humans to reliably annotate factuality.
To resolve this issue, we further define \textsc{Advanced ReviewScore} as an aggregation of premise factuality scores.
If a weakness has no supporting reason, we call it as a \textit{claim}, and if a weakness consists of more than one premises, we call it as an \textit{argument}.
We adopt two aggregation methods, logical conjunction, following the literature of logic~\citep{sep-logical-consequence}, and weighted average, maintaining a 5-point scale (Figure \ref{fig:main}).
Before evaluating \textsc{Advanced ReviewScore}, we have to extract all \textit{explicit} and \textit{implicit premises} from an argument. This process is called \textit{argument reconstruction} in logic and critical thinking~\citep{gregor-argrecon, informal-logic, dowden}, and we construct an engine that automatically does this process (Section \ref{subsec:auto-argrecon}).
To evaluate the ability of LLMs for detecting \textit{misinformed} review points, we construct a human expert-annotated \textsc{ReviewScore} dataset based on ICLR reviews (Section \ref{subsec:dataset}). We recruited 15 experienced graduate students and they dedicated total 244 hours for trustworthy human annotation.

We examine the reliability of automatic evaluation of \textsc{ReviewScore} by measuring human-model agreements (Section \ref{sec:experiments}). We use eight current state-of-the-art LLMs, including five proprietary and three open-sourced models, and the results show that there are only moderate agreements with the human experts (i.e., 0.4--0.5 F1 Scores and 0.3--0.4 Kappa Scores). A thorough human-model disagreement analysis reveals that models sometimes misunderstand the meaning of review points or predict scores that minority human annotators give.
Furthermore, \textsc{Advanced ReviewScore} clearly outperforms \textsc{Base ReviewScore}, which proves the effectiveness of premise-level factuality scoring.

To summarize our contributions:
\begin{enumerate}[leftmargin=7mm]
\vspace{-2mm}
    \item We introduce \textsc{ReviewScore}, a novel evaluation criteria that detects \textit{misinformed} review points (i.e., questions that can be answered by the paper or weaknesses with incorrect premises).
    \item To evaluate the factuality of premises, we propose an automatic argument reconstruction engine that generates a \textit{valid} and \textit{faithful} set of premises and conclusion.
    \item We construct a trustworthy human expert-annotated dataset to measure the reliability of automatic evaluation of \textsc{ReviewScore}.
    \item We validate that an automatic \textsc{ReviewScore} evaluation with current state-of-the-art LLMs only shows moderate agreement with human experts and conduct a comprehensive human-model disagreement analysis.
\vspace{-2mm}
\end{enumerate}

\section{Related Works}
\label{sec:related}
\paragraph{Peer review evaluation.}
Previous works studied how to evaluate the quality of peer reviews.
\citet{shin2025mind} and \citet{purkayastha} map reviews to a predefined set of review types (i.e., types of facets and lazy thinking, respectively). However, they do not cover groundedness or factuality of reviews, which are emphasized as \textit{understanding} and \textit{substantiation} in \citet{goldberg}.
\citet{guo}, \citet{sadallah}, and \citet{claimcheck} evaluate reviews by deciding whether the reviews are grounded and supported by a target paper or mapping reviews to a set of claims of the paper. The downside is that they do not explicitly evaluate whether the reviews are factually correct or not, leaving it as a future work.
\citet{review-critique} go one step further and evaluate factuality of every sentence in weaknesses. However, the main drawback is that a sentence-level factuality cannot fully capture correctness of an underlying logic of weaknesses.
Our work resolves this issue by first reconstructing a weakness into a set of explicit and implicit premises and then evaluating factuality of each premise. Moreover, our work also explicitly evaluates a quality of questions, whereas the aforementioned prior works mostly focus on evaluating weaknesses or a full review.
These comparisons are summarized in Table~\ref{tab:eval_dimensions}.

\paragraph{Argument evaluation.}
In logic and critical thinking, an argument is a list of statements, one of which is the conclusion and the others are the premises~\citep{sep-argument, argument-def}.
To evaluate an argument, we need to follow two steps. First, we have to identify and reconstruct the argument into a set of premises and conclusion, which is called an argument reconstruction~\citep{gregor-argrecon, informal-logic, dowden}. Then, we evaluate whether each premise is factually correct.
An argument reconstruction should both be \textit{valid}, which means premises deductively imply a conclusion, and \textit{faithful}, which means premises and a conclusion accurately and completely represents an original argument~\citep{gregor-argrecon, deepa2}.
Previously, \citet{deepa2} trains a T5 model for argument reconstruction. However, the training datasets are either synthetic or polished and the reconstruction do not require any additional context information.
In contrast, our work targets peer reviews, which include real-world unpolished arguments, and the reconstruction requires an entire paper to fully understand the context of arguments.

\section{ReviewScore}
\label{sec:method}
We newly define \textsc{ReviewScore} that measures how \textit{misinformed} a review point is (Section~\ref{subsec:reviewscore-def}). To evaluate \textsc{ReviewScore}, we also introduce an engine that automatically extracts every \textit{explicit} and \textit{implicit premises} from a weakness (Section~\ref{subsec:auto-argrecon}). Lastly, we construct a human expert-annotated dataset that evaluates LLM's ability to evaluate \textsc{ReviewScore} (Section~\ref{subsec:dataset}).

\subsection{Definition}
\label{subsec:reviewscore-def}
Our goal of defining \textsc{ReviewScore} is to detect \textit{misinformed} review points.
Following this goal and the review quality criteria discussion in Section \ref{sec:intro}, we first define \textsc{Base ReviewScore} as \textit{factuality of weaknesses} and \textit{unanswerability of questions}. The following definition formally describes it.
\begin{definition}[\textsc{Base ReviewScore}]
Let $x$ be a review point (either a weakness or a question) about a submitted paper $S$. Define
\[
\begin{aligned}
\operatorname{Factuality}_S &:\mathcal{W}\to\{1,2,3,4,5\},\\
\operatorname{Unanswerability}_S &:\mathcal{Q}\to\{1,2,3,4,5\},
\end{aligned}
\]
where $\mathcal{W}$ and $\mathcal{Q}$ are, respectively, the sets of weaknesses and questions appearing in a review of $S$\footnote{Detailed rubric is described in Appendix \ref{sec:prompts}.}.
We considered score 1--2 as \textit{Misinformed} and score 3--5 as \textit{Not misinformed} for binary classification setup.
The \textsc{Base ReviewScore} of $x$ is:
\[
\begin{split}
\operatorname{ReviewScore}_{\mathrm{base}}(x)~~~~~~~~~~~~~~~~~~~~~~~~~~~~~~~~~\\
=\begin{cases}
\operatorname{Factuality}_S(x) & \text{if } x\in\mathcal{W},\\
\operatorname{Unanswerability}_S(x) & \text{if } x\in\mathcal{Q}.
\end{cases}
\end{split}
\]
For simplicity, we call the \textsc{Base ReviewScore} of weakness and questions as WScore and QScore.
\end{definition}

However, during the group discussion in Section \ref{sec:intro}, we discovered that human annotators' evaluations on factuality diverge if a weakness contains both factual and nonfactual premises or a nonfactual premise is implicitly presumed.
It happens since the human annotators implicitly weigh the importance of underlying premises of a weakness and then decide the final factuality score.

To resolve this issue, we further define \textsc{Advanced ReviewScore}.
We categorize weaknesses into arguments and claims based on whether there are supporting reasons or not. We keep the definition of WScore to evaluate claims, but further develops a finer-grained score to evaluate arguments. Following the literature of critical thinking~\citep{gregor-argrecon, informal-logic, dowden}, we reconstruct an argument into a premise-conclusion structure and then define \textsc{Advanced ReviewScore} for arguments as (an aggregation of) \textit{factuality of premises}. The following definition formally describes it.
\begin{definition}[\textsc{Advanced ReviewScore}]
Let $x$ be a review point about a submitted paper $S$. Let $\mathcal{C}$, $\mathcal{A}$, and $\mathcal{Q}$ denote, respectively, the sets of claims, arguments, and questions in a review of $S$.  
For $x\in\mathcal{A}$, let its (\textit{explicit} and \textit{implicit}) premises be
$\mathcal{P}(x)=\{p_1,\dots,p_k\}$ with $\{p_1,\dots,p_k\}\vdash C$ for the conclusion $C$ of $x$.
Let $\mathcal{K}$ be the set of knowledge bases available for factuality judgments (i.e., $S$, annotator knowledge, referred papers), and let
\[
\begin{aligned}
\operatorname{Factuality}&:\ \mathcal{U}\times\mathcal{K}\to\{1,2,3,4,5\},\\
\mathcal{U}&\coloneqq \mathcal{C}\ \cup\ \bigcup_{x\in\mathcal{A}}\mathcal{P}(x),
\end{aligned}
\]
be a 5-point scoring function for claims and premises given a knowledge base.
Define a selector $\mathrm{KB}:\mathcal{U}\to\mathcal{K}$ that chooses the knowledge base used
for each item (for claims $x\in\mathcal{C}$, $\mathrm{KB}(x)=S$; for premises $p_i$,
$\mathrm{KB}(p_i)=\mathrm{KB}_i\in\mathcal{K}$).  
For a given $\mathrm{KB}(x)\in\mathcal{K}$, we notate the factuality function as $\operatorname{Factuality}_{\mathrm{KB}(x)}(\cdot)$.
$\mathrm{Agg}$ is an operator that aggregates a list of scores to a single score. The \textsc{Advanced ReviewScore} of $x$ is:
\[
\begin{split}
\operatorname{ReviewScore}_{\mathrm{adv}}(x)~~~~~~~~~~~~~~~~~~~~~~~~~~~~~~~~~~~~~~~~~~~~~~~~~~\\=
\begin{cases}
\operatorname{Factuality}_{S}(x)
& \text{if } x\in\mathcal{C}\ ,\\[8pt]
\mathrm{Agg}\!\left(\operatorname{Factuality}_{\mathrm{KB}_i}(p_i)_{p_i\in\mathcal{P}(x)}\right)
& \text{if } x\in\mathcal{A}\ ,\\[12pt]
\operatorname{Unanswerability}_{S}(x)
& \text{if } x\in\mathcal{Q}\ .
\end{cases}
\end{split}
\]
For simplicity, we call the \textsc{Advanced ReviewScore} of claims, arguments, and questions as ClaimScore, ArgScore, and QScore.
\end{definition}

\paragraph{Aggregation methods.}
To aggregate premise factuality scores as a single ArgScore, we introduce two aggregation methods: logical conjunction and weighted average.
Following the literature of logic, an argument is \textit{true} if and only if all premises are \textit{true}.
We define a premise is \textit{true} if and only if it has factuality score 3--5, and otherwise, it is \textit{false}.
We dubbed this aggregation as logical conjunction, which follows the binary classification setup.
However, if an annotator mistakenly evaluates one of premises, then the error propagates to the entire argument.
To alleviate this issue, we also aggregate by a weighted average. Since it is difficult to measure the importance of premises, we instead weighted scores by untrivialness of premises (with a scale of 0--2). This is intended to simply filter out trivially true premises by measuring their importance as 0.\footnote{Since we reconstruct every argument as valid (i.e., a set of premises deductively implies a conclusion), there are often conditional premises that make an argument valid but are trivially true. Detailed rubric of untrivialness is described in Appendix \ref{sec:prompts}.} Specifically, (un)trivialness is decided based on the common knowledge of CS/AI-majoring undergrad students.

\subsection{Automatic Argument Reconstruction}
\label{subsec:auto-argrecon}

\begin{figure*}
  \vspace{-6mm}
  \centering

  \begin{subfigure}[t]{\linewidth}
    \centering
    \includegraphics[width=\linewidth]{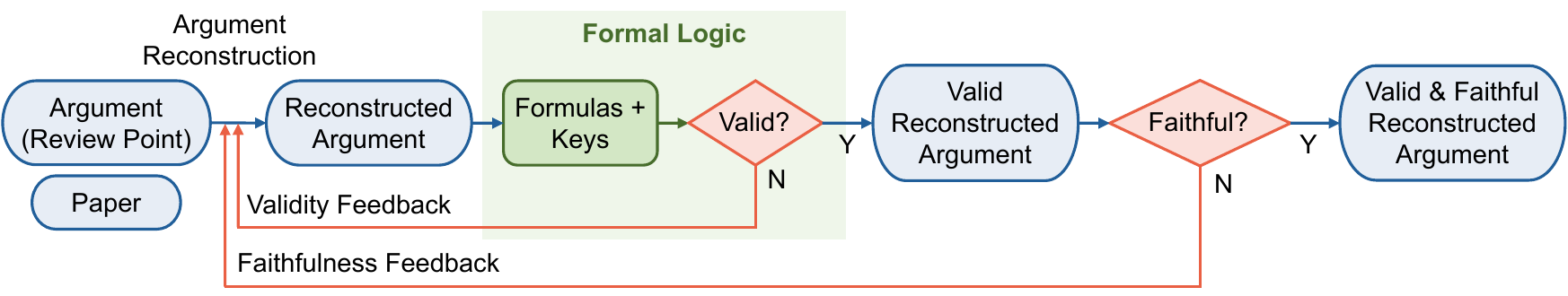}
    \caption{}
    \label{fig:argrecon-overview}
  \end{subfigure}


  \begin{subfigure}[t]{\linewidth}
    \centering
    \includegraphics[width=\linewidth]{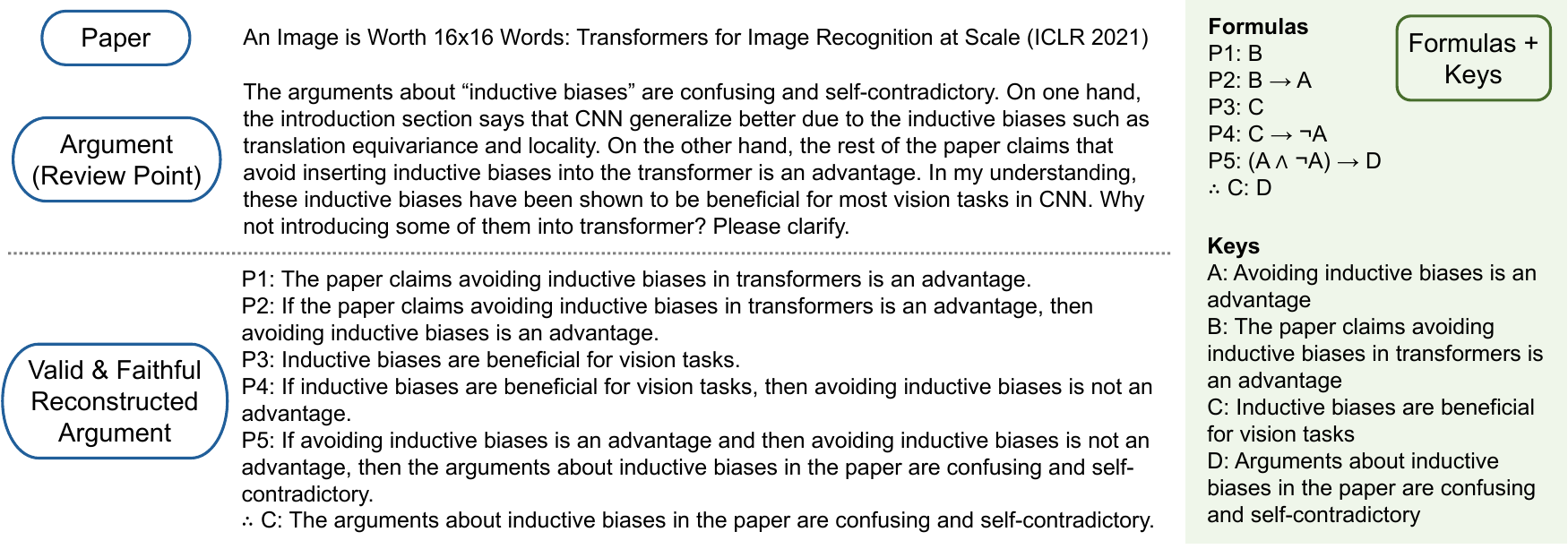}
    \caption{}
    \label{fig:argrecon-ex}
  \end{subfigure}
  \caption{
  (a) Overview of an automatic argument reconstruction.
  Given an argumentative review point with a paper, a model first generates a reconstructed argument (i.e., a set of premises and conclusion).
  To check its \textit{validity}, a model translates a NL reconstructed argument into FOL formulas, and then a SAT solver judges if it is \textit{valid}.
  To check its \textit{faithfulness}, a model translate FOL formulas back into the NL domain, and a model judges if the reconstruction is \textit{faithful}.
  If one of two criteria does not met, then corresponding NL feedback is given to the generator model.
  (b) A representative example.
  We sample a review point of \cite{vit} and its reconstruction along with corresponding formulas and keys.
  }
  \label{fig:argrecon}
  \vspace{-5mm}
\end{figure*}

To evaluate ArgScore, we have to extract (\textit{explicit} and \textit{implicit}) premises $\mathcal{P}(x)$ from an argument $x\in\mathcal{A}$. Since human experts require significant amount of time and costs to do this, we automate it using LLMs. 
First, we check if a model could directly reconstruct an argument by giving detailed instructions. To preserve the context of an argument, we also give the model a submitted paper $S$. However, it mostly fails to generate \textit{valid} and \textit{faithful} reconstructions\footnote{Detailed numerical results and qualitative analysis are reported in Appendix \ref{sec:add-arg-recon}.}.
To resolve this issue, we add two feedback loops to ensure \textit{validity} and \textit{faithfulness} of an argument reconstruction (Figure \ref{fig:argrecon-overview}).

An LLM alone often falls short in ensuring \textit{validity}, so we include a SAT solver which could automatically judges the \textit{validity} of a set of premises and a conclusion without any logical errors.
To do that, an LLM translates a set of natural language (NL) premises and conclusion into first-order logic (FOL) formulas. Then, a SAT solver decides whether the premises deductively implies the conclusion.
If the reconstruction is \textit{valid}, then an LLM translates formalized premises and conclusion back into NL domain. This process is called \textit{logical streamlining}, which means to rephrase NL premises or conclusion in order to make their logico-semantic structure more transparent~\citep{critical-thinking, gregor-argrecon, deepa2}. We then pass these \textit{streamlined} NL premises and conclusion to the subsequent faithfulness feedback loop.
However, if the reconstruction is \textit{invalid}, then we feed a rule-based reward signal to the argument reconstructor to regenerate the reconstruction. There are two types of reward signals, one is a naive signal that says the formalized premises do not imply the conclusion, and the other one tells that the proof is circular.

Although the reconstruction is now \textit{valid}, it should also \textit{faithfully} represents the original argument.
To ensure that, we prompt an LLM to judge whether the reconstruction is \textit{faithful} or not and justify its decision.
If the reconstruction is \textit{faithful}, then we stop iterating the loop.
If the reconstruction is \textit{unfaithful}, then we feed LLM judge's justification to the argument reconstructor to regenerate the reconstruction.
To minimize model calls in practice, we connect these two loops in series so that only \textit{valid} reconstructions are judged for their \textit{faithfulness}.
Also, we limit the total number of loop iteration to 10 and return the last reconstructed argument if the loop fails to generate \textit{valid} and \textit{faithful} reconstruction.
We provide an example reconstruction in Figure \ref{fig:argrecon-ex}, and details on feedback loops and model prompts are described in Appendix \ref{sec:add-arg-recon} and \ref{sec:prompts}, respectively.

\paragraph{Quality of argument reconstruction.}
We measure \textit{validity} and \textit{faithfulness} of reconstructed arguments using a SAT solver and human annotators, respectively. With Claude Sonnet 3.7 as a base LLM, every reconstruction is \textit{valid}, and average \textit{faithfulness} score is 4.5 / 5.
Detailed analysis including comparison with direct reconstruction is described in Appendix \ref{sec:add-arg-recon}.

\subsection{Dataset Construction}
\label{subsec:dataset}
Our final destination of proposing \textsc{ReviewScore} is to \textit{automatically} filter out \textit{misinformed} review points using LLMs. To verify LLM's ability to do that, we build a human expert-annotated dataset to measure an agreement between humans and LLMs on \textsc{ReviewScore} evaluation.
Our dataset contains total 657 annotated review points, consisting of 143 questions, 92 claims, and 422 arguments. Specifically, 1,748 premises of the arguments are manually annotated
\footnote{Details of the dataset construction and statistics are described in Appendix \ref{sec:add-dataset}.}.

\paragraph{Human-review matching.}
We recruit 15 graduate students studying AI as human annotators, and they annotate total 40 papers submitted to ICLR 2021--2023\footnote{Since ICLR 2024--2025 submission drafts are not opened to public, we exclude these years.}.
Specifically, we first make five groups by their research interests. For each group, three human annotators discuss which papers to annotate and select eight papers that are relevant with all three. Then, each human annotates selected eight papers' review points which are preprocessed from OpenReview.

\paragraph{Data curation process.}
We collaboratively use an LLM and humans for \textsc{ReviewScore} data curation, where an LLM preprocesses reviews and then humans annotate those.
Given a review, an LLM extracts independent review points.
For each review point, an LLM automatically annotates the type (i.e., claim, argument, or question) and human verifies it.
If the review point is a question, then human scores if the question is (un)answerable by the paper in a 5-point scale and justifies it if needed.
If the review point is a claim, then human scores if the claim is true in a 5-point scale and justifies it if needed.
If the review point is an argument, then human scores the argument's factuality same as in claims (i.e., \textsc{Base ReviewScore}).
To annotate \textsc{Advanced ReviewScore} for arguments, we first run automatic argument reconstruction engine (Section \ref{subsec:auto-argrecon}) to extract underlying premises of the argument.
After that, human scores if the reconstruction is \textit{faithful} in a 5-point scale. If the \textit{faithfulness} score is less than 4 (i.e., faithful, but one or two minor changes recommended), then they skip the subsequent annotations. Otherwise, they judge the factuality of premises.
For each premise, they first select a knowledge base (i.e., submitted paper, annotator knowledge, or referred papers), score if the premise is true based on the knowledge base in a 5-point scale, and score if the premise is (un)trivial in a 3-point scale. They justify any of three decisions if needed.

\paragraph{Trustworthiness of human annotation.}
We provide annotators detailed guidelines and an hour-long online orientation session. During annotation, they are allowed to use any related materials or tools (e.g., discussion between authors and reviewers in OpenReview, web search, etc.).
Furthermore, we highly encourage the annotators to communicate with us through a group chat so that we could give them instant responses to their questions and share with all, which builds a global consensus among human annotators.

To reduce annotator bias, three humans first independently annotate review points. To control label quality, we further conduct a disagreement-aware recheck where humans are given disagreed instances\footnote{Annotations are disagreed if a difference of maximum and minimum scores is larger than or equal to 2.} and revise these only if clear mistakes are indicated. Note that humans independently perform this recheck without having access to others' labels.
Thanks to these efforts, despite the difficulty of the work, we obtain median 0.489 and highest 0.663 inter-annotator agreement in Krippendorff's Alpha~\citep{krippendorffsalpha} across annotator groups.

However, there still exists few disagreements after the recheck. To resolve these, we conduct an additional group discussion to finalize human labels of the disagreed instances.
We basically take median values from three annotations as final human labels, whereas for few disagreed instances, we exceptionally take post-discussion labels as final ones.
Details are described in Appendix \ref{sec:analysis-human}.

\section{Reliability of Automatic Evaluation of ReviewScore}
\label{sec:experiments}
To evaluate LLM's ability to evaluate \textsc{ReviewScore}, we describe an experimental setup (Section~\ref{subsec:setup}) and show human-model agreement results (Section~\ref{subsec:main-results}). We also compare the effectiveness of \textsc{Base} and \textsc{Advanced ReviewScore} (Section~\ref{subsec:base-vs-advanced}). We further analyze human-model disagreements and the effect of providing authors response to models (Section~\ref{subsec:analysis}).

\begin{table*}[t]
\vspace{-5mm}
\caption{
Human-model agreement on \textsc{ReviewScore} evaluation.
}
\label{table:main}
\centering
\small
\setlength{\tabcolsep}{6pt}
\begin{tabularx}{0.8\linewidth}{lcccccccc}
\toprule
& \multicolumn{2}{c}{ClaimScore} & \multicolumn{2}{c}{ArgScore} & \multicolumn{2}{c}{QScore} & \multicolumn{2}{c}{\textsc{ReviewScore}} \\
\cmidrule(lr){2-3}\cmidrule(lr){4-5}\cmidrule(lr){6-7}\cmidrule(lr){8-9}
Model & F1 & Kappa & F1 & Kappa & F1 & Kappa & F1 & Kappa \\
\midrule
\textit{Proprietary models} \\
Claude Sonnet 3.7   & 0.160 & 0.156 & 0.462 & 0.261 & 0.576 & 0.410 & 0.462 & 0.336 \\
Claude Sonnet 4     & 0.222 & 0.165 & 0.403 & 0.272 & \textbf{0.579} & \textbf{0.425} & 0.448 & \textbf{0.378} \\
GPT-4o              & 0.000 & 0.093 & 0.333 & 0.244 & 0.476 & 0.291 & 0.385 & 0.347 \\
GPT-5               & 0.125 & 0.024 & \textbf{0.481} & 0.353 & 0.560 & 0.340 & \textbf{0.482} & 0.341 \\
Gemini 2.5 Flash    & \textbf{0.240} & \textbf{0.169} & 0.466 & \textbf{0.402} & 0.522 & 0.265 & 0.464 & 0.357 \\
\midrule
\textit{Open-sourced models} \\
Qwen3-235B-A22B     & \textbf{0.231} & 0.142 & \textbf{0.413} & 0.148 & \textbf{0.544} & 0.243 & \textbf{0.452} & 0.262 \\
Llama 3.3           & 0.118 & 0.097 & 0.338 & 0.108 & 0.514 & \textbf{0.254} & 0.415 & \textbf{0.311} \\
DeepSeek-V3         & 0.000 & \textbf{0.180} & 0.298 & \textbf{0.176} & 0.530 & 0.192 & 0.383 & 0.301 \\
\bottomrule
\end{tabularx}
\vspace{-5mm}
\end{table*}

\subsection{Setup}
\label{subsec:setup}
Given score rubrics in a 5-point scale, an LLM evaluates \textsc{ReviewScore} according to a knowledge base it selects. We only provide a submitted paper to a model since we assume that the model has a human-level internal knowledge and has a general understanding of referred papers. Detailed model prompts are described in Appendix \ref{sec:prompts}.

\paragraph{Language models.}
To measure LLM's ability to automatically evaluate \textsc{ReviewScore}, we perform experiments on eight current state-of-the-art LLMs that achieve significantly high alignments with humans. We include five proprietary models, Claude Sonnet 3.7~\citep{claude-3-7-sonnet}, Claude Sonnet 4~\citep{claude-sonnet-4}, GPT-4o~\citep{gpt4o}, GPT-5~\citep{gpt5}, and Gemini 2.5 Flash~\citep{gemini}, and three open-sourced models, Qwen3-235B-A22B~\citep{qwen3}, DeepSeek-V3~\citep{deepseek-v3}, and Llama 3.3~\citep{llama3}\footnote{
Details of model specifications are described in Appendix \ref{sec:model-details}.
}.
To get consistent and reliable scores from LLM judges, we set a low temperature (i.e., 0) and select the highest probability response from the model~\citep{Liang2022HELM, Liu2023GEval, Gu2024LLMJudgeSurvey}.
We exclude reasoning models as LLM judges since \textsc{ReviewScore} mostly depends on grounding and evidence, not longer or smarter chains of thought.

\paragraph{Evaluation metrics.}
We use different sets of metrics for two types of problem formulations, binary classification and 5-point scale scoring.
For the binary classification, since majority of human-annotated labels are \textit{Not misinformed}, we mainly use F1 score which is robust to class imbalance.
For the 5-point scale scoring, since majority of human-annotated scores are 4 and 5, we mainly use Quadratic Weighted Kappa~\citep{qwk}, a variant of Cohen's Kappa~\citep{cohens}, that is robust to the skewed data distribution. Hereinafter, we call this metric Kappa for simplicity.
To provide more comprehensive results, we additionally use Precision and Recall for the binary classification and Pearson rank correlation and Gwet's AC2~\citep{gwet} for the 5-point scale scoring.

\subsection{Main Results}
\label{subsec:main-results}

We empirically validate the alignment of human-annotated and model-estimated \textsc{ReviewScore} using different models and evaluation metrics in Table \ref{table:main}.
Most models show 0.4--0.5 F1 score and 0.3--0.4 Kappa score, which indicates moderate agreement between humans and models on \textsc{ReviewScore} evaluation.
However, there are differences in human-model agreements for three types of review points. Regardless of the models, questions show the highest agreement, arguments follow subsequently, and claims show the lowest agreement. Specifically, for claims, some models show zero F1 score or nearly zero Kappa score. We analyze human-model disagreements thoroughly in Section \ref{subsec:analysis} and conclude that since claims lack supporting evidence and are often value-laden, models often misinterpret the intended meaning of the claims or judge differently than humans.
More results and qualitative analysis of model evaluation are described in Appendix \ref{sec:add-exps}.

\begin{table*}[t]
\vspace{-5mm}
\caption{
Comparison of human-model agreement of \textsc{Base} vs. \textsc{Advanced ReviewScore}.
}
\label{table:base-adv}
\centering
\small
\setlength{\tabcolsep}{6pt}
\begin{tabularx}{0.63\linewidth}{llccc}
\toprule
Model & Metric & Base & Advanced w/o Agg & Advanced \\
\midrule
\multirow{2}{*}{Claude Sonnet 3.7} & F1      & 0.157 & 0.333 & \textbf{0.462} \\
                                           & Kappa   & 0.120 & \textbf{0.308} & 0.261 \\
\multirow{2}{*}{GPT-5}             & F1      & 0.218 & 0.370 & \textbf{0.481} \\
                                           & Kappa   & 0.182 & \textbf{0.366} & 0.353 \\
\multirow{2}{*}{Gemini 2.5 Flash}  & F1      & 0.137 & 0.395 & \textbf{0.466} \\
                                           & Kappa   & 0.124 & 0.386 & \textbf{0.402} \\
\multirow{2}{*}{DeepSeek-V3}       & F1      & 0.146 & 0.167 & \textbf{0.298} \\
                                           & Kappa   & 0.091 & \textbf{0.193} & 0.176 \\
\bottomrule
\end{tabularx}
\vspace{-3mm}
\end{table*}

\subsection{Base vs. Advanced ReviewScore}
\label{subsec:base-vs-advanced}

We compare human-model agreements of \textsc{Base} and \textsc{Advanced ReviewScore} using different models and evaluation metrics. Since the only difference is in defining scores for arguments, we compare \textsc{ReviewScore} for arguments in Table \ref{table:base-adv}. To verify the effectiveness of aggregation methods, we additionally include \textsc{Advanced ReviewScore} for arguments without aggregation (i.e., factuality scores of premises).
Regardless of models, \textsc{Advanced ReviewScore} clearly shows higher agreements than \textsc{Base ReviewScore}. For Gemini 2.5 Flash, \textsc{Advanced ReviewScore} performs 3.40$\times$ higher F1 score and 3.24$\times$ higher Kappa score than the \textsc{Base ReviewScore}.
For the aggregation methods, logical conjunction contributes to higher agreements (i.e., F1 scores are consistently higher with aggregation).
However, weighted average shows marginal improvements or slight degradations. We manually investigate human-model disagreements and observe that humans and models have slightly misaligned criteria on whether a given premise is trivially true or not based on the common knowledge of CS/AI-majoring undergrad students.

\begin{figure}[t]
  \centering
  \includegraphics[width=\columnwidth]{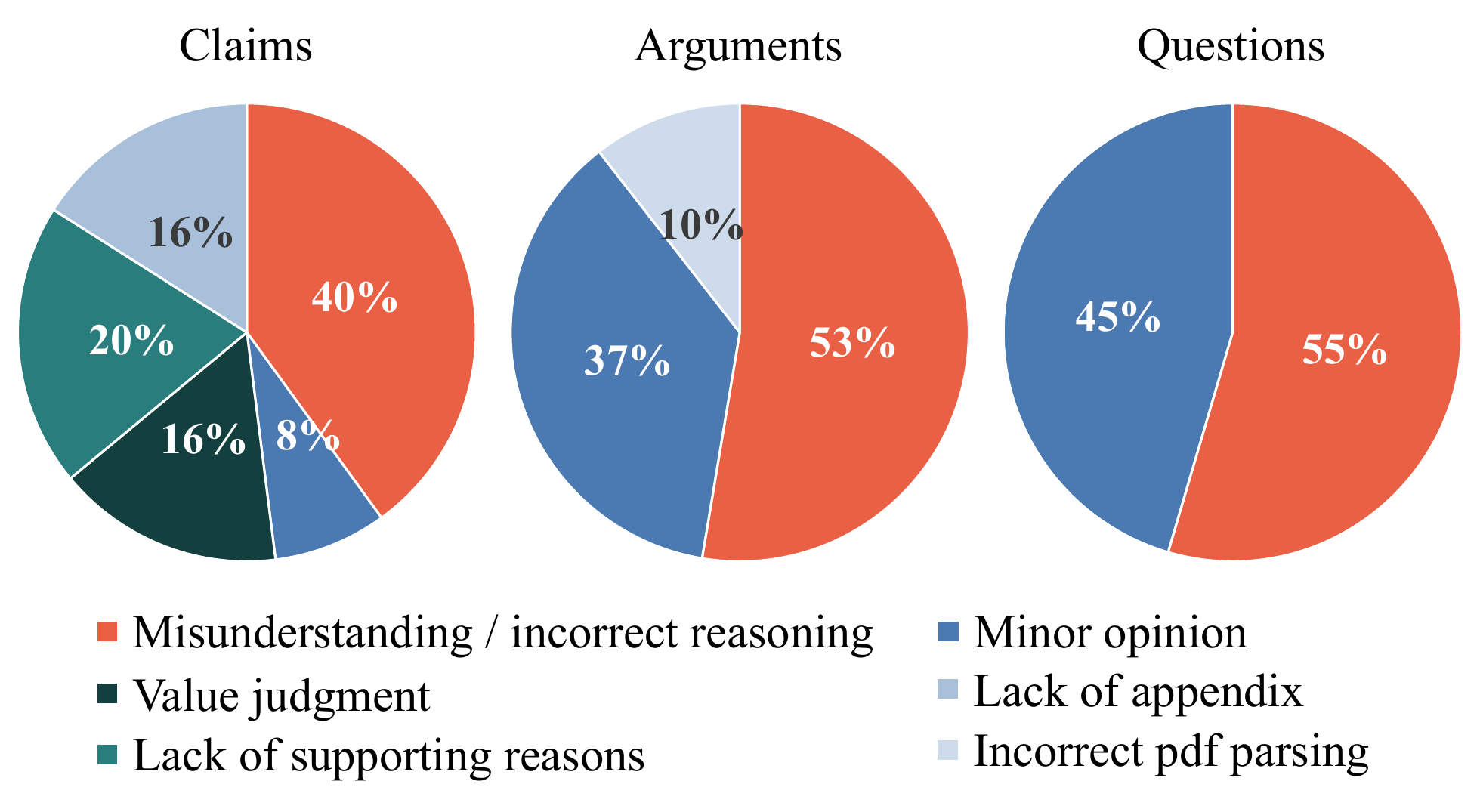}
  \caption{
  Types of human-model disagreements.
  }
  \label{fig:error_review}
  \vspace{-5mm}
\end{figure}

\subsection{Analysis}
\label{subsec:analysis}

\paragraph{Human-model disagreements.}
We analyze types of human-model disagreements in Figure \ref{fig:error_review}.
Across all review points, there are two common types of disagreements, which are models' misunderstanding or incorrect reasoning and models' predictions which correspond to minority of human-annotated scores. These two types comprise nearly all disagreements on arguments and questions, and about half of disagreements on claims.
For claims, we observe that 36\% of disagreements are caused by either claims are value-laden or lack supporting reason. These make the factuality judgment of claims subjective, leading to a low human-model agreement.
Practical limitations such as incorrect pdf parsing and not providing appendix to models cause nontrivial portion of disagreements.

\paragraph{Helpfulness of authors response.}
We study whether providing authors responses of reviews to a model benefits the automatic \textsc{ReviewScore} evaluation using Claude Sonnet 4 in Table \ref{table:author_response}.
Overall, providing authors response leads to higher human-model agreement of \textsc{ReviewScore} evaluation.
However, we observe that providing authors response has marginal or sometimes negative effect on QScore and ClaimScore evaluation.
These observations show that authors response largely benefits argument or premise factuality evaluation by providing additional cues for model judgments, whereas does not benefit question (un)answerability and claim factuality by injecting authors bias.

\begin{table}[t]
\caption{
Effect of providing Authors Response (AR) to a model for \textsc{ReviewScore} evaluation.
}
\label{table:author_response}
\centering
\small
\setlength{\tabcolsep}{4pt}
\begin{tabularx}{\columnwidth}{lcccc}
\toprule
& \multicolumn{2}{c}{F1} & \multicolumn{2}{c}{Kappa}\\
\cmidrule(lr){2-3}\cmidrule(lr){4-5}
 & w/o AR & w/ AR & w/o AR & w/ AR \\
\midrule
WScore           & 0.194 & \textbf{0.243} & 0.149 & \textbf{0.198} \\
ArgScore w/o Agg & 0.315 & \textbf{0.385} & 0.322 & \textbf{0.372} \\
\midrule
ArgScore         & 0.403 & \textbf{0.493} & 0.272 & \textbf{0.338} \\
QScore           & \textbf{0.579} & \textbf{0.578} & \textbf{0.425} & 0.410 \\
ClaimScore       & \textbf{0.222} & 0.125 & 0.165 & \textbf{0.221} \\
\midrule
\textsc{ReviewScore}     & 0.448 & \textbf{0.498} & 0.378 & \textbf{0.416} \\
\bottomrule
\end{tabularx}
\vspace{-5mm}
\end{table}

\section{Conclusion}
\label{sec:conclusion}
We introduce \textsc{ReviewScore}, a new evaluation metric of peer review quality, focusing on detecting \textit{misinformed} review points.
To evaluate LLM's ability for \textsc{ReviewScore} evaluation, we also construct a trustworthy human-annotated dataset. The results show a moderate human-model agreement, and further comprehensive disagreement analysis reveals that current state-of-the-art LLMs sometimes misunderstand or reason incorrectly.
However, we confirm that premise-level factuality shows significantly higher human-model agreements than weakness-level factuality, which proves the effectiveness of our method.

\section*{Limitations}
We acknowledge three types of limitations of our work.

First, there are practical limitations in the automated \textsc{ReviewScore} evaluation.
To save API calling costs, when the paper is given to LLMs, we only provide text and tables of the main paper, where any figure or appendix is excluded.
We also observe nontrivial amount of incorrect pdf parsing.

Second, there are technical limitations in the automatic argument reconstruction.
The reconstruction output is not always perfect since it depends on the capability of the base model. However, we observe a considerable output quality improvement when we upgrade the base model to Claude Sonnet 4, indicating that the reconstruction engine would perform better as the model improves in general.

Lastly, there are practical limitations in the dataset construction.
Since human annotators are graduate students with varying skillfulness and the annotation requires significant amount of cognitive load, there are unavoidable noise in human annotation.
Furthermore, a manual selection of papers by human annotators might introduce any kind of unintended biases. However, to collect the most reliable annotation under our limited budget, we inevitably choose this method.

\section*{Acknowledgments}

We would like to thank Ji Yong Cho, Carolyn Rose, Sean Welleck, and Nathaniel Weir for providing insightful comments and discussions.

\bibliography{custom}

\clearpage

\appendix

\section{Future Impact}
\label{sec:future}
We expect automatic \textsc{ReviewScore} evaluation could greatly benefit different roles of the current peer reviewing system.
For authors, by providing reconstruction of argumentative review points, it helps to understand or clarify the reviewer's intention and to formulate a rebuttal.
For reviewers, by providing \textsc{ReviewScore} of their review points, it allows them to verify the review quality by themselves and helps reviewers to better understand the paper.
For metareviewers, by providing \textsc{ReviewScore} of each reviewer, it could assist their final decisions.
To summarize, it could serve as an automated system for managing the review quality.

\section{Review Quality Criteria Discussion}
\label{sec:criteria-selection}
First, we recruit a group of three graduate students studying AI and NLP and let them independently analyze reviews of ten common manually selected submitted papers from ICLR 2021--2023\footnote{Since those papers include papers written by authors of our work, we do not share the full list of target papers to keep anonymity. We will uncover the full list after the paper is published.}. Specifically, we guided the group to: decompose a review into several independent review points, evaluate quality of the review points by a 5-point scale with their own criteria, and justify their scores that ends with a meta-sentence that does not involve paper's context which is used for our further analysis. During human evaluation, we minimize our effort to provide detailed guidelines to facilitate bias-free human analysis.

Then, we categorize these justifications in order to find common features of low-quality review points.
There are five common types: (1) questions that can be already addressed by the paper, (2) comments that reflect a misunderstanding of the paper, (3) out-of-scope remarks, (4) observations pointing out minor details, and (5) unclear points.
However, for the last three types, we observe that majority of review points are agreed or argued by a single human annotator. In other words, given a review point, only a single human argues it is out-of-scope, whereas other two humans argue it is within scope and could be a potential drawback of the submitted paper. Following this pattern, a single human argues a review point is addressing minor details but others argue it is a major point, and a single human argues a review point is unclear whereas others do not agree with this.
In contrast, the first two types of review points (i.e., answerable questions and misunderstood comments) are mostly agreed by two or more human annotators, indicating that these two types have more objective and trustworthy criteria for detecting low-quality reviews. Based on these observations, we focus on evaluating review points based on the first two criteria.

Lastly, we share some meaningful insights from this group discussion that strongly motivates our work.
Based on the analysis of score justifications, given a review point, some (sub)sentences are high-quality or factually correct but others could be misinformed or factually incorrect. This later motivates our premise-level factuality evaluation (Section \ref{subsec:reviewscore-def}).
Furthermore, during the group discussion, we observe that human annotators sometimes struggle which parts in a review point they should weigh more to evaluate the review point. This later motivates our aggregation methods which include logical conjunction and weighted average by untrivialness (Section \ref{subsec:reviewscore-def}).
We also discover that human evaluations are sometimes incorrect. We leave this as a limitation of our work.
However, to alleviate this issue, we recruit three human annotators for every instance and further ensure the annotation quality by providing them careful guidelines and actively communicating with them to build a global consensus on the evaluation criteria (Section \ref{subsec:dataset}).

\section{Automatic Argument Reconstruction Details and Results}
\label{sec:add-arg-recon}
\subsection{Implementation details}
We elaborate components of automatic argument reconstruction engine (Section \ref{subsec:auto-argrecon}, Figure \ref{fig:argrecon}) in the following. We also refer the corresponding prompt used in each step if it exists.
\begin{enumerate}[leftmargin=7mm]
    \item Given an argumentative review point, an LLM extracts a verbatim conjecture and its verbatim reason statements (Figure \ref{fig:prompt_recon_verbatim}).
    \item Given a verbatim conclusion and reasons of the argumentative review point and a corresponding paper parsed from a pdf file, an LLM reconstructs the argument into a premise-conclusion structure. At the same inference, the model also translates (or formalizes) NL premises and conclusion into corresponding FOL formulas and generate keys which assign NL meaning to variables and predicates. To facilitate the model to generate a \textit{valid} formalization, the model generates a deductive proof using formalized premises and conclusion at the end (Figure \ref{fig:prompt_recon_recon}).
    \item Given FOL premises and conclusion with keys and a deductive proof, an LLM extracts necessary FOL premises for the deductive proof, write a python program using Z3~\citep{z3} that automatically checks the \textit{validity} of the necessary FOL premises and conclusion, extracts a final FOL conclusion that is used in the python program, and judges whether the proof is circular (i.e., whether the final FOL conclusion is included in one of necessary FOL premises.) (Figure \ref{fig:prompt_recon_validity}, Figure \ref{fig:prompt_recon_validity_code}).
    \item If the proof is circular, then a NL feedback indicating circularity of the proof is sent to Step 2, and the model re-generates an argument reconstruction. Otherwise, we run the python program that checks the \textit{validity} of the reconstruction. However, if the program returns an error, the model takes this error message and re-generate the python program that fixes the error (Figure \ref{fig:prompt_recon_debug}).
    If the reconstruction is \textit{invalid}, then a NL feedback indicating \textit{invalidity} of the reconstruction is sent to Step 2, and the model re-generates an argument reconstruction. Otherwise (i.e., if the reconstruction is \textit{valid}), we proceed to the next step.
    \item To check \textit{faithfulness} of the reconstruction, the model first translates FOL premises and conclusion (one of the outputs of the model in Step 2) with keys (one of the outputs of the model in Step 1) into NL premises and conclusion (Figure \ref{fig:prompt_recon_deformalization}). This process is called \textit{logical streamlining} in logic and critical thinking~\citep{critical-thinking, gregor-argrecon, deepa2}.
    \item Lastly, given an original argumentative review point (or an argument) and the \textit{streamlined} NL premises and conclusion, the model judges whether the reconstruction is \textit{faithful} with justifications (Figure \ref{fig:prompt_recon_faithfulness}). If the reconstruction is \textit{unfaithful}, then a NL feedback including the model's justifications is sent to Step 2, and the model re-generates an argument reconstruction accordingly. Otherwise, since the reconstruction is \textit{valid} yet \textit{faithful}, the feedback loop is finished and the \textit{streamlined} NL premises and conclusion become a final argument reconstruction.
\end{enumerate}

\subsection{Quantitative Results}

To verify effectiveness of the feedback loop, we report the argument reconstruction performance with and without feedback in Table \ref{table:argrecon_perf}.
We mainly measure average \textit{validity} and \textit{faithfulness} of the reconstructed arguments using a SAT solver and human annotators, respectively. We provide score rubric for evaluating \textit{faithfulness} in Figure \ref{fig:rubric_recon_faithfulness}.
We also include the pass rate, which indicates whether the (last) reconstructed argument fulfill the \textit{validity} and \textit{faithfulness} criteria in the feedback loop.
Furthermore, we report the average number of loop iterations to check if the feedback loops are actively used.

In Table \ref{table:argrecon_perf}, we verify that the proposed method (i.e., w/ feedback) achieves a perfect \textit{validity} and nearly perfect \textit{faithfulness} and pass rate, whereas the performance of the direct reconstruction (i.e., w/o feedback) largely lags behind that.
We observe that the average number of loop iterations is 3.09, indicating active usage of feedback signals, but the number varies a lot depending on the clarity of logical structures of the arguments.

\begin{table}[h]
\caption{Performance of Automatic Argument Reconstruction.}
\label{table:argrecon_perf}
\centering
\small
\setlength{\tabcolsep}{6pt}
\begin{tabularx}{0.8\linewidth}{l c c}
\toprule
 & w/o feedback & w/ feedback \\
\midrule
Validity     & 0.895 & \textbf{1.00} \\
Faithfulness & 2.91   & \textbf{4.47} \\
Pass Rate    & 0.369 & \textbf{0.935} \\
\midrule
\# of loops   & N/A   & 3.09 $\pm$ 2.73 \\
\bottomrule
\end{tabularx}
\end{table}

\begin{figure}[h]
\footnotesize

\begin{tcblisting}{text only,
    halign=left, 
    title=\textbf{Faithfulness Rubric for Argument Reconstruction},
    colbacktitle=gray!30!white, 
    coltitle=black,
}
5: Definitely faithful, no change is essential\\
4: Faithful, but one or two minor details in premises need to change in order to fully expres s the original context\\
3: Faithful, but more than two minor details in premises need to change\\
2: Not faithful, one or two major error exist\\
1: Definitely not faithful, more than two major errors exist
\end{tcblisting}
\vspace{-2mm}
\caption{Score rubric for evaluating \textit{faithfulness} of argument reconstruction by human annotators.}
\label{fig:rubric_recon_faithfulness}
%
\end{figure}

\subsection{Qualitative Results}

We present three representative examples of model-generated argument reconstruction and compare outputs with and without the feedback loop.
We observe the direct reconstruction (i.e., w/o feedback) often generates either \textit{invalid} or \textit{unfaithful} reconstructed arguments, however, the proposed method (i.e., w/ feedback) generates \textit{valid} yet \textit{faithful} reconstructed arguments.
We provide detailed analysis in Figure \ref{fig:error_analysis_1}--\ref{fig:error_analysis_3}.

\begin{figure*}[!h]
\footnotesize
\begin{tcblisting}{text only,
    halign=left, 
    title=\textbf{Input: Argument (Review Point) \& Paper}, 
    colbacktitle=gray!30!white, 
    coltitle=black
}
\vspace{5pt}

\textit{Paper}:\\
Least-to-Most Prompting Enables Complex Reasoning in Large Language Models~\citep{g5p3}

\hdashrule{\linewidth}{0.4pt}{.5mm} \vspace{-0pt}
\textit{Argument (Review Point)}:\\
The paper lacks insights on key questions. Does the model have ability to decompose questions very different from demo examples into correct sub-problems? Even the model solves each sub-problem correctly, what makes the model compose the final solution correctly?

\vspace{5pt}
\end{tcblisting}

\begin{tcblisting}{text only,
    halign=left, 
    title=\textbf{Output: Reconstructed Argument}, 
    colbacktitle=gray!30!white, 
    coltitle=black
}
\vspace{5pt}

\textit{Direct Reconstruction (w/o feedback)}:\\
P1: The paper does not demonstrate whether the model has the ability to decompose questions very different from demo examples into correct sub-problems.\\
P2: The paper does not explain what makes the model compose the final solution correctly even when it solves each sub-problem correctly.\\
P3: If \textcolor{red}{a paper fails to provide insights on key questions about a proposed method}, then the paper lacks insights on key questions.\\
P4: Questions about decomposition ability and solution composition are key questions for the least-to-most prompting method.\\
$\therefore$~C: The paper lacks insights on key questions.

\hdashrule{\linewidth}{0.4pt}{.5mm} \vspace{-0pt}
\textit{Proposed Method (w/ feedback)}:\\
P1: The paper does not demonstrate whether the model has the ability to decompose questions very different from demo examples into correct sub-problems.\\
P2: If the paper does not demonstrate whether the model has the ability to decompose questions very different from demo examples into correct sub-problems, then the paper fails to provide insights on the key question about decomposition ability.\\
P3: If the paper fails to provide insights on the key question about decomposition ability or fails to provide insights on the key question about solution composition, then the paper fails to provide insights on key questions.\\
P4: If the paper fails to provide insights on key questions, then the paper lacks insights on key questions.\\
$\therefore$~C: The paper lacks insights on key questions.

\vspace{5pt}
\end{tcblisting}

\begin{tcblisting}{text only,
    halign=left, 
    title=\textbf{Analysis}, 
    colbacktitle=gray!30!white, 
    coltitle=black
}
\vspace{5pt}
~- For the direct reconstruction, P1, P2, P4 semantically implies an intermediate conclusion that ``The paper fails to provide insights on key questions about the proposed method.'', but this connecting premise is not explicitly reconstructed. Therefore, this reconstruction is \textcolor{red}{\textit{invalid}}~[\xmark].
However, if we consider that this connecting premise is implicitly presumed, then the reconstruction \textit{faithfully}~[\cmark]~represents the original argument. \\
~- For the proposed method, the reconstruction is logically \textit{valid}~[\cmark]~yet \textit{faithfully}~[\cmark]~represents the original argument.
\vspace{5pt}
\end{tcblisting}
\vspace{-2mm}
\caption{Example \#1 of automatic argument reconstruction.}
\label{fig:error_analysis_1}
\end{figure*}

\begin{figure*}[!h]
\footnotesize
\begin{tcblisting}{text only,
    halign=left, 
    title=\textbf{Input: Argument (Review Point) \& Paper}, 
    colbacktitle=gray!30!white, 
    coltitle=black
}
\vspace{5pt}

\textit{Paper}:\\
ReAct: Synergizing Reasoning and Acting in Language Models~\citep{g5p2}

\hdashrule{\linewidth}{0.4pt}{.5mm} \vspace{-0pt}
\textit{Argument (Review Point)}:\\
ReAct alone works fairly well for some tasks, but self-consistent Chain of Thought often outperforms ReAct alone, and leads to fairly dramatic increases in performance. The best models use a smart combination of both CoT and ReAct, using heuristics that likely maximize performance on the dataset(s) (see `ReAct $\rightarrow$ CoT--SC' strategy), which makes the performance of ReAct by itself slightly less impressive (though I still think this is a useful contribution for the community)

\vspace{5pt}
\end{tcblisting}

\begin{tcblisting}{text only,
    halign=left, 
    title=\textbf{Output: Reconstructed Argument}, 
    colbacktitle=gray!30!white, 
    coltitle=black
}
\vspace{5pt}

\textit{Direct Reconstruction (w/o feedback)}:\\
P1: The best models use a smart combination of both CoT and ReAct.\\
P2: Models use heuristics that likely maximize performance on the dataset(s) (see `ReAct $\rightarrow$ CoT--SC' strategy).\\
P3: If models use heuristics that likely maximize performance on the dataset(s), then the performance of ReAct by itself is less impressive.\\
P4: If the best models use a combination of both CoT and ReAct rather than ReAct alone, then the performance of ReAct by itself is less impressive.\\
P5: \textcolor{red}{The performance of ReAct by itself being less impressive still means it is a useful contribution for the community.}\\
$\therefore$~C: \textcolor{red}{ReAct alone works fairly well for some tasks, but self-consistent Chain of Thought often outperforms ReAct alone, and leads to fairly dramatic increases in performance.}

\hdashrule{\linewidth}{0.4pt}{.5mm} \vspace{-0pt}
\textit{Proposed Method (w/ feedback)}:\\
P1: ReAct alone works fairly well for some tasks.\\
P2: Self-consistent Chain of Thought often outperforms ReAct alone.\\
P3: The best models use a smart combination of both CoT and ReAct.\\
P4: Models use heuristics like the `ReAct $\rightarrow$ CoT--SC' strategy that maximize performance.\\
P5: If Self-consistent Chain of Thought often outperforms ReAct alone, and the best models use a smart combination of both CoT and ReAct, and models use heuristics like the `ReAct $\rightarrow$ CoT--SC' strategy that maximize performance, then the performance of ReAct by itself is slightly less impressive.\\
P6: If ReAct alone works fairly well for some tasks, and the performance of ReAct by itself is slightly less impressive, and a technique can be a useful contribution even if its performance is less impressive, then ReAct is still a useful contribution for the community.\\
P7: A technique can be a useful contribution even if its performance is less impressive.\\
$\therefore$~C: The performance of ReAct by itself is slightly less impressive, and ReAct is still a useful contribution for the community.

\vspace{5pt}
\end{tcblisting}

\begin{tcblisting}{text only,
    halign=left, 
    title=\textbf{Analysis}, 
    colbacktitle=gray!30!white, 
    coltitle=black
}
\vspace{5pt}
~- For the direct reconstruction, both pairs of premises, P1 \& P4, and P2 \& P3, deductively imply an intermediate conclusion that ``The performance of ReAct by itself is less impressive.''. However, there is no logical connection between this intermediate conclusion and the final conclusion C, which indicates the reconstruction is \textcolor{red}{\textit{invalid}}~[\xmark]. Furthermore, the final conclusion C does not correctly represent the original argument, which means the reconstruction is \textcolor{red}{\textit{unfaithful}}~[\xmark].\\
~- For the proposed method, the premises deductively imply the final conclusion (i.e., \textit{valid}~[\cmark]), and the reconstruction correctly yet completely represents the original argument (i.e., \textit{faithful}~[\cmark]), including the correct final conclusion.
\vspace{5pt}
\end{tcblisting}
\vspace{-2mm}
\caption{Example \#2 of automatic argument reconstruction.}
\label{fig:error_analysis_2}
\end{figure*}

\begin{figure*}[!h]
\footnotesize
\begin{tcblisting}{text only,
    halign=left, 
    title=\textbf{Input: Argument (Review Point) \& Paper}, 
    colbacktitle=gray!30!white, 
    coltitle=black
}
\vspace{5pt}

\textit{Paper}:\\
Automatic Chain of Thought Prompting in Large Language Models~\citep{g5p1}

\hdashrule{\linewidth}{0.4pt}{.5mm} \vspace{-0pt}
\textit{Argument (Review Point)}:\\
Finally, if clustering was the way to go, the authors could have tried to cluster samples into $n$-clusters, see where the test sample is fitting, and could have tried to use just one prompt instead of $k$ different ones at each time. This might have reduced the compute cost and would be interesting to see the performance vs resource trade-off.

\vspace{5pt}
\end{tcblisting}

\begin{tcblisting}{text only,
    halign=left, 
    title=\textbf{Output: Reconstructed Argument}, 
    colbacktitle=gray!30!white, 
    coltitle=black
}
\vspace{5pt}

\textit{Direct Reconstruction (w/o feedback)}:\\
P1: The authors could have tried to cluster samples into n-clusters and see where the test sample is fitting.\\
P2: The authors could have tried to use just one prompt instead of $k$ different ones at each time.\\
P3: Using one prompt instead of $k$ different ones would reduce the compute cost.\\
P4: If a method reduces compute cost, it would be interesting to see the performance vs resource trade-off.\\
P5: If the authors could have done something that would be interesting to see, then they should have done it.\\
$\therefore$~C: \textcolor{red}{The authors should have tried to cluster samples into $n$-clusters, see where the test sample is fitting, and use just one prompt instead of $k$ different ones at each time.}\\

\hdashrule{\linewidth}{0.4pt}{.5mm} \vspace{-0pt}
\textit{Proposed Method (w/ feedback)}:\\
P1: An alternative approach would be to cluster samples into $n$-clusters, see where the test sample is fitting, and use just one prompt instead of $k$ different ones.\\
P2: If an alternative approach would use just one prompt instead of $k$ different ones, then the alternative approach would reduce compute cost.\\
P3: If the alternative approach would reduce compute cost, then examining the performance vs. resource trade-off of the alternative approach would be interesting.\\
P4: If an alternative approach would use just one prompt instead of $k$ different ones and examining the performance vs. resource trade-off of the alternative approach would be interesting, then the alternative approach would have been interesting to evaluate for its performance vs. resource trade-off.\\
$\therefore$~C: The alternative approach would have been interesting to evaluate for its performance vs. resource trade-off.

\vspace{5pt}
\end{tcblisting}

\begin{tcblisting}{text only,
    halign=left, 
    title=\textbf{Analysis}, 
    colbacktitle=gray!30!white, 
    coltitle=black
}
\vspace{5pt}
~- For the direct reconstruction, P3 and P4 imply an intermediate conclusion that ``Using one prompt instead of k different ones would be interesting to see the performance vs resource trade-off.''. Let us call this IC1. IC1 and P5 imply another intermediate conclusion that ``The authors should have tried using one prompt instead of k different ones.''. Let us call this IC2. Lastly, P1, P2, and IC2 imply the final conclusion C, indicating the reconstruction is \textit{valid}~[\cmark].
However, the final conclusion is stronger than what the original argument states. The original argument suggests an alternative approach, but the reconstructed final conclusion obligates authors to try it, meaning the reconstruction is \textcolor{red}{\textit{unfaithful}}~[\xmark].\\
~- For the proposed method, the reconstruction is \textit{valid}~[\cmark]~and \textit{faithful}~[\cmark], including the correct degree of strength of the final conclusion.
\vspace{5pt}
\end{tcblisting}
\vspace{-2mm}
\caption{Example \#3 of automatic argument reconstruction.}
\label{fig:error_analysis_3}
\end{figure*}

\clearpage

\section{Dataset Details}
\label{sec:add-dataset}
\subsection{Terms of Use and License}
All papers and corresponding reviews used in our work are crawled from OpenReview\footnote{\url{https://openreview.net}}, and our work is consistent with the Openreview terms of use\footnote{\url{https://openreview.net/legal/terms}}.
Following these terms, we will release our work with a CC-BY 4.0 license.

\enlargethispage{2\baselineskip}

\subsection{Pilot Dataset}
\label{subsec:pilot-setup}
Before we construct the main dataset in Section \ref{subsec:dataset}, to ensure a trustworthiness and reliability of human annotation, we conduct a pilot study.
We recruit three graduate students studying AI as human annotators, and let them choose total five papers submitted to ICLR 2021--2025\footnote{Unlike the main dataset, we include ICLR 2024--2025 since authors can access to their own submissions.}.
Specifically, each human annotates reviews of three papers in OpenReview, where only one paper is authored by themselves and the other two non-authored papers are assigned in common.
To ensure trustworthiness of non-authors' annotation, three humans annotate reviews and then we take a median value as a final human label.

\subsection{Dataset Statistics}
We present the \textsc{ReviewScore} dataset statistics including the pilot and main subsets in Table \ref{table:dataset-stats}. We include number of instances and percentage of \textit{misinformed} labels.

\begin{table}[h]
\caption{\textsc{ReviewScore} dataset statistics.}
\label{table:dataset-stats}
\centering
\small
\setlength{\tabcolsep}{6pt}
\begin{tabularx}{0.93\linewidth}{l c c c}
\toprule
 & Pilot & Main & Total \\
\midrule
ICLR Years & 2021--2025 & 2021--2023 & - \\
\# Papers & 5 & 40 & 45 \\
\# Review(er)s & 19 & 155 & 174 \\
\midrule
\multicolumn{4}{c}{\textit{Number of instances}} \\
Review Points & 84 & 573 & 657 \\
\hspace*{1em}Questions & 22 & 121 & 143 \\
\hspace*{1em}Claims & 5 & 87 & 92 \\
\hspace*{1em}Arguments & 57 & 365 & 422 \\
\hspace*{2em}Premises & 227 & 1{,}521 & 1{,}748 \\
\bottomrule
\end{tabularx}
\end{table}

\begin{table*}[t]
\caption{
Full list of ICLR submitted papers used in the \textsc{ReviewScore} dataset.
}
\label{table:full-paper}
\centering
\small
\setlength{\tabcolsep}{6pt}
\begin{tabularx}{0.9\linewidth}{ccl}
\toprule
Group & Topic & Paper Title \\
\midrule
\multirow{3}{*}{1}& \multirow{3}{*}{Image Generation} & \citet{g1p1}, \citet{g1p2}, \citet{g1p3}, \\
                  && \citet{g1p4}, \citet{g1p5}, \citet{g1p6}, \citet{g1p7}, \\
                  && \citet{g1p8} \\
\midrule
\multirow{3}{*}{2}& \multirow{3}{*}{Time Series ML} & \citet{g2p1}, \citet{g2p2}, \citet{g2p3}, \\
                  && \citet{g2p4}, \citet{g2p5}, \citet{g2p6}, \citet{g2p7}, \\
                  && \citet{g2p8} \\
\midrule
\multirow{2}{*}{3}& LLM Reasoning / & \citet{g3p1}, \citet{g3p2}, \citet{g3p3}, \citet{g3p4}, \\
                  & Compression & \citet{g3p5}, \citet{g3p6}, \citet{g3p7}, \citet{g3p8}\\
\midrule
\multirow{2}{*}{4}& \multirow{2}{*}{LLM / VLM} & \citet{g4p1}, \citet{g4p2}, \citet{g4p3}, \citet{g4p4}, \\
                  &  & \citet{g4p5}, \citet{g4p6}, \citet{g4p7}, \citet{g4p8}\\
\midrule
\multirow{3}{*}{5}& \multirow{3}{*}{LLM Prompting} & \citet{g5p1}, \citet{g5p2}, \citet{g5p3}, \\
                  &  & \citet{g5p4}, \citet{g5p5}, \citet{g5p6}, \\
                  && \citet{g5p7}, \citet{g5p8} \\
\bottomrule
\end{tabularx}
\end{table*}

\subsection{Full List of Papers}
We provide a full list of papers used in the main subset of \textsc{ReviewScore} dataset in Table \ref{table:full-paper}. As explained in Section \ref{subsec:dataset}, eight papers are selected by each human annotator group with a common research interest.

\section{Analysis of Human Annotators}
\label{sec:analysis-human}
\subsection{Human Annotator Expertise}
\label{subsec:annotator-analysis}

We report human annotator's expertise on \textsc{ReviewScore} evaluation in Table \ref{table:annotator-info}. Specifically, for each annotator, we indicate a number of publications in AI/ML (including arXived works) and an averaged paper relevance score across eight assigned papers. We also indicate averaged numbers and scores for each group.
The results show that an average number of publication is 3.93 and average paper relevance score is 4.06 / 5, indicating highly-experienced and relevant experts conduct a human annotation process.
However, we observe that there are inter-group gaps in human expertise. Specifically, Group 2 and 4 shows significantly higher number of publications and paper relevance than other groups.
Detailed score rubric for paper relevance is described in Figure \ref{fig:rubric_paper_rel}, and we note that there is no instance where paper relevance of any human annotator is less than 3 (i.e., Moderate relevance).

\begin{table*}[h]
\caption{
Human annotator's expertise on \textsc{ReviewScore} evaluation. A \textbf{bold} indicates the highest number/score across different groups, and an \underline{underline} indicates the second highest.
}
\label{table:annotator-info}
\centering
\small
\setlength{\tabcolsep}{6pt}
\begin{tabularx}{0.8\linewidth}{ccccccc}
\toprule
Group & Annotator ID & \# AI/ML Pub & Paper Relevance & Avg \# Pub & Avg Paper Relevance \\
\midrule
\multirow{3}{*}{1}& anno\_11 & 2 & 3.50 & \multirow{3}{*}{3.00} & \multirow{3}{*}{4.00} \\
                  & anno\_12 & 3 & 4.13 && \\
                  & anno\_13 & 4 & 4.38 && \\
\midrule
\multirow{3}{*}{2}& anno\_21 & 6 & 4.50 & \multirow{3}{*}{\underline{5.33}} & \multirow{3}{*}{\textbf{4.38}} \\
                  & anno\_22 & 3 & 4.38 && \\
                  & anno\_23 & 7 & 4.25 && \\
\midrule
\multirow{3}{*}{3}& anno\_31 & 0 & 4.00 & \multirow{3}{*}{1.33} & \multirow{3}{*}{3.96} \\
                  & anno\_32 & 1 & 4.00 && \\
                  & anno\_33 & 3 & 3.88 && \\
\midrule
\multirow{3}{*}{4}& anno\_41 & 1 & 4.00 & \multirow{3}{*}{\textbf{6.67}} & \multirow{3}{*}{\underline{4.04}} \\
                  & anno\_42 & 6 & 3.88 && \\
                  & anno\_43 & 13 & 4.25 && \\
\midrule
\multirow{3}{*}{5}& anno\_51 & 3 & 3.13 & \multirow{3}{*}{3.33} & \multirow{3}{*}{3.92} \\
                  & anno\_52 & 1 & 4.13 && \\
                  & anno\_53 & 6 & 4.50 && \\
\midrule
Total & - & - & - & 3.93 & 4.06 \\
\bottomrule
\end{tabularx}
\end{table*}

\begin{figure}[h]
\footnotesize

\begin{tcblisting}{text only,
    halign=left, 
    title=\textbf{Paper Relevance Score Rubric},
    colbacktitle=gray!30!white, 
    coltitle=black,
}
5: Direct expertise -- Works in the exact subtopic and can judge nuanced claims, methods, and datasets.\\
4: Strong relevance -- Adjacent/overlapping subtopic with regular use of the paper’s methods or domain; can evaluate technical choices with minimal ramp-up.\\
3: Moderate relevance -- Same broad area (e.g., NLP $\leftrightarrow$ NLP; CV $\leftrightarrow$ CV) but different subtopic or methods; will understand contributions but may miss edge-case nuances.\\
2: Low relevance -- Only tangential connection (e.g., general ML experience while the paper is domain-specific) and limited familiarity with core methods or domain.\\
1: No clear relevance -- Outside the field; would require substantial background reading to assess claims/methodology.
\end{tcblisting}
\vspace{-2mm}
\caption{Score rubric for evaluating paper relevance of human annotators.}
\label{fig:rubric_paper_rel}
%
\end{figure}

\subsection{Inter-Annotator Agreement}
\label{subsec:inter-annotator}
To ensure trustworthiness of human annotation, we report inter-annotator agreement in Krippendorff's Alpha~\citep{krippendorffsalpha} on \textsc{ReviewScore} evaluation in Table \ref{table:inter-annotator}.
Overall, \textsc{ReviewScore} shows 0.489 Krippendorff's Alpha, indicating a moderate inter-annotator agreement. Specifically, QScore shows the highest agreement, ArgScore follows subsequently, and ClaimScore shows the lowest agreement.

By comparing inter-group agreements, Group 2 and 3 show significantly higher \textsc{ReviewScore} agreement than other groups.
In contrast, Group 1 and 5 show lower agreement than other groups. Through a manual disagreement analysis, we confirm that most disagreements come from human annotators with low paper relevance (i.e., anno\_11 and anno\_51 in Table \ref{table:annotator-info}).
This means that human annotations could be more reliable if their research interests become more relevant to the assigned papers. We leave this as a limitation of our work.

\begin{table*}[h]
\caption{
Inter-annotator agreement (Krippendorff's Alpha) on \textsc{ReviewScore} evaluation.
A \textbf{bold} indicates the highest agreement across different groups, and an \underline{underline} indicates the second highest.
}
\label{table:inter-annotator}
\centering
\small
\setlength{\tabcolsep}{6pt}
\begin{tabularx}{0.55\linewidth}{ccccc}
\toprule
Group & ClaimScore & ArgScore & QScore & \textsc{ReviewScore} \\
\midrule
1 & 0.392 & 0.339 & 0.357 & 0.357 \\
2 & \underline{0.438} & \textbf{0.600} & \textbf{0.780} & \textbf{0.663} \\
3 & \textbf{0.660} & 0.457 & \underline{0.655} & \underline{0.580} \\
4 & 0.356 & \underline{0.535} & 0.554 & 0.489 \\
5 & 0.235 & 0.306 & 0.335 & 0.254 \\
\midrule
Median & 0.392 & 0.457 & 0.554 & 0.489 \\
\bottomrule
\end{tabularx}
\end{table*}

\section{Additional Results of Automatic ReviewScore Evaluation}
\label{sec:add-exps}
\subsection{Quantitative Results}
\label{subsec:add-quantitative-results}

We additionally report human-model agreement on \textsc{ReviewScore} evaluation using different evaluation metrics in Table \ref{table:claimscore}--\ref{table:reviewscore}. Specifically, we report agreement on ClaimScore evaluation in Table \ref{table:claimscore}, agreement on ArgScore evaluation in Table \ref{table:argscore}, agreement on QScore evaluation in Table \ref{table:qscore}, and agreement on \textsc{ReviewScore} evaluation in Table \ref{table:reviewscore}.
For the binary classification setup, we use Precision, Recall, and F1 Score, and for the 5-point scale setup, we use Pearson rank correlation, Gwet's AC2~\citep{gwet}, and Quadratic Weighted Kappa~\citep{qwk}.

\begin{table*}[h]
\caption{
Human-model agreement on ClaimScore evaluation.
}
\label{table:claimscore}
\centering
\small
\setlength{\tabcolsep}{6pt}
\begin{tabularx}{0.7\linewidth}{lcccccc}
\toprule
& \multicolumn{3}{c}{Binary} & \multicolumn{3}{c}{5-point Scale} \\
\cmidrule(lr){2-4}\cmidrule(lr){5-7}
Model & Precision & Recall & F1 & Pearson & AC2 & Kappa \\
\midrule
\textit{Proprietary models} \\
Claude Sonnet 3.7   & 0.091 & \textbf{0.667} & 0.160 & 0.217 & 0.056 & 0.156 \\
Claude Sonnet 4     & \textbf{0.167} & 0.333 & 0.222 & 0.215 & 0.086 & 0.165 \\
GPT-4o              & 0.000 & 0.000 & 0.000 & 0.098 & 0.036 & 0.093 \\
GPT-5               & 0.091 & 0.200 & 0.125 & 0.020 & -0.028 & 0.024 \\
Gemini 2.5 Flash    & 0.158 & 0.500 & \textbf{0.240} & \textbf{0.226} & \textbf{0.119} & \textbf{0.169} \\
\midrule
\textit{Open-sourced models} \\
Qwen3-235B-A22B     & \textbf{0.143} & \textbf{0.600} & \textbf{0.231} & 0.191 & 0.096 & 0.142 \\
Llama 3.3           & 0.083 & 0.200 & 0.118 & 0.083 & 0.063 & 0.097 \\
DeepSeek-V3         & 0.000 & 0.000 & 0.000 & \textbf{0.213} & \textbf{0.102} & \textbf{0.180} \\
\bottomrule
\end{tabularx}
\end{table*}
\begin{table*}[h]
\caption{
Human-model agreement on ArgScore evaluation.
}
\label{table:argscore}
\centering
\small
\setlength{\tabcolsep}{6pt}
\begin{tabularx}{0.7\linewidth}{lcccccc}
\toprule
& \multicolumn{3}{c}{Binary} & \multicolumn{3}{c}{5-point Scale} \\
\cmidrule(lr){2-4}\cmidrule(lr){5-7}
Model & Precision & Recall & F1 & Pearson & AC2 & Kappa \\
\midrule
\textit{Proprietary models} \\
Claude Sonnet 3.7   & 0.357 & 0.652 & 0.462 & 0.323 & 0.238 & 0.261 \\
Claude Sonnet 4     & 0.313 & 0.565 & 0.403 & 0.296 & 0.231 & 0.272 \\
GPT-4o              & \textbf{0.550} & 0.239 & 0.333 & 0.230 & 0.237 & 0.244 \\
GPT-5               & 0.419 & 0.565 & \textbf{0.481} & 0.401 & 0.342 & 0.353 \\
Gemini 2.5 Flash    & 0.356 & \textbf{0.674} & 0.466 & \textbf{0.469} & \textbf{0.387} & \textbf{0.402} \\
\midrule
\textit{Open-sourced models} \\
Qwen3-235B-A22B     & 0.413 & \textbf{0.413} & \textbf{0.413} & \textbf{0.295} & -0.016 & 0.148 \\
Llama 3.3           & \textbf{0.480} & 0.261 & 0.338 & -0.002 & 0.110 & 0.108 \\
DeepSeek-V3         & 0.292 & 0.304 & 0.298 & 0.181 & \textbf{0.155} & \textbf{0.176} \\
\bottomrule
\end{tabularx}
\end{table*}
\begin{table*}[h]
\caption{
Human-model agreement on QScore evaluation.
}
\label{table:qscore}
\centering
\small
\setlength{\tabcolsep}{6pt}
\begin{tabularx}{0.7\linewidth}{lcccccc}
\toprule
& \multicolumn{3}{c}{Binary} & \multicolumn{3}{c}{5-point Scale} \\
\cmidrule(lr){2-4}\cmidrule(lr){5-7}
Model & Precision & Recall & F1 & Pearson & AC2 & Kappa \\
\midrule
\textit{Proprietary models} \\
Claude Sonnet 3.7   & \textbf{0.576} & 0.576 & 0.576 & 0.416 & 0.404 & 0.410 \\
Claude Sonnet 4     & 0.524 & 0.647 & \textbf{0.579} & \textbf{0.460} & \textbf{0.406} & \textbf{0.425} \\
GPT-4o              & 0.392 & 0.606 & 0.476 & 0.336 & 0.247 & 0.291 \\
GPT-5               & 0.512 & 0.618 & 0.560 & 0.362 & 0.326 & 0.340 \\
Gemini 2.5 Flash    & 0.393 & \textbf{0.774} & 0.522 & 0.330 & 0.201 & 0.265 \\
\midrule
\textit{Open-sourced models} \\
Qwen3-235B-A22B     & 0.400 & \textbf{0.848} & \textbf{0.544} & 0.331 & 0.132 & 0.243 \\
Llama 3.3           & 0.380 & 0.794 & 0.514 & \textbf{0.335} & \textbf{0.167} & \textbf{0.254} \\
DeepSeek-V3         & \textbf{0.415} & 0.733 & 0.530 & 0.277 & 0.067 & 0.192 \\
\bottomrule
\end{tabularx}
\end{table*}
\begin{table*}[h]
\caption{
Human-model agreement on \textsc{ReviewScore} evaluation.
}
\label{table:reviewscore}
\centering
\small
\setlength{\tabcolsep}{6pt}
\begin{tabularx}{0.7\linewidth}{lcccccc}
\toprule
& \multicolumn{3}{c}{Binary} & \multicolumn{3}{c}{5-point Scale} \\
\cmidrule(lr){2-4}\cmidrule(lr){5-7}
Model & Precision & Recall & F1 & Pearson & AC2 & Kappa \\
\midrule
\textit{Proprietary models} \\
Claude Sonnet 3.7   & 0.367 & 0.622 & 0.462 & 0.368 & 0.315 & 0.336 \\
Claude Sonnet 4     & 0.365 & 0.581 & 0.448 & \textbf{0.414} & \textbf{0.349} & \textbf{0.378} \\
GPT-4o              & 0.392 & 0.378 & 0.385 & 0.351 & 0.340 & 0.347 \\
GPT-5               & \textbf{0.421} & 0.565 & \textbf{0.482} & 0.362 & 0.327 & 0.341 \\
Gemini 2.5 Flash    & 0.347 & \textbf{0.699} & 0.464 & \textbf{0.415} & \textbf{0.349} & \textbf{0.378} \\
\midrule
\textit{Open-sourced models} \\
Qwen3-235B-A22B     & 0.365 & \textbf{0.595} & \textbf{0.452} & \textbf{0.351} & 0.178 & 0.262 \\
Llama 3.3           & \textbf{0.370} & 0.471 & 0.415 & 0.012 & \textbf{0.299} & \textbf{0.311} \\
DeepSeek-V3         & 0.327 & 0.462 & 0.383 & 0.338 & 0.259 & 0.301 \\
\bottomrule
\end{tabularx}
\end{table*}

\subsection{Qualitative Results}
\label{subsec:add-qualitative-results}

We present a thorough qualitative human-model disagreement analysis on \textsc{ReviewScore} evaluation in Figure \ref{fig:error_analysis_claim_1}--\ref{fig:error_analysis_q_2}.
For each review point type, we demonstrate two examples which include model-based scores and corresponding justifications using Claude Sonnet 3.7 and Claude Sonnet 4.

\begin{figure*}[!h]
\footnotesize
\begin{tcblisting}{text only,
    halign=left, 
    title=\textbf{Input: Claim (Review Point) \& Paper}, 
    colbacktitle=gray!30!white, 
    coltitle=black
}
\vspace{5pt}

\textit{Paper}:\\
Towards Continual Knowledge Learning of Language Models~\citep{g5p6}

\hdashrule{\linewidth}{0.4pt}{.5mm} \vspace{-0pt}
\textit{Claim (Review Point)}:\\
The experimental findings in this paper are somewhat trivial.

\vspace{5pt}
\end{tcblisting}

\begin{tcblisting}{text only,
    halign=left, 
    title=\textbf{Human-annotated Label (5-point Scale)}, 
    colbacktitle=gray!30!white, 
    coltitle=black
}
\vspace{5pt}
3, 4, 4 (Final: 4)
\vspace{5pt}
\end{tcblisting}

\begin{tcblisting}{text only,
    halign=left, 
    title=\textbf{Output: ClaimScore Evaluation},
    colbacktitle=gray!30!white, 
    coltitle=black
}
\vspace{5pt}

\textit{Model}: Claude Sonnet 3.7\\
~\\

\textit{Score}: \textcolor{red}{\textbf{2}}\\
~\\

\textit{Human-model Disagreement Type}: \textcolor{red}{Value judgment}\\
~\\

\textit{Justification}: The proposition states that the experimental findings in this paper are somewhat trivial. To evaluate this claim, I need to assess the novelty, significance, and depth of the experimental findings presented in the paper.\\

Key experimental findings in the paper include:\\

\hspace{6.2cm} $\dots$

These findings are not trivial because:\\
- They establish a new problem formulation and benchmark for an important real-world challenge\\
- They provide novel insights about how different methods perform in this setting\\
- They identify key factors affecting knowledge retention and acquisition\\
- The findings have practical implications for developing ever-changing language models\\
% ~\\

% The experimental design is thorough, with multiple scenarios tested (single-phase vs. multi-phase CKL, different corpus sizes, etc.) and detailed analyses of the results.

\hdashrule{\linewidth}{0.4pt}{.5mm} \vspace{-0pt}
\textit{Model}: Claude Sonnet 4\\
~\\

\textit{Score}: \textcolor{red}{\textbf{2}}\\
~\\

\textit{Human-model Disagreement Type}: \textcolor{red}{Value judgment}\\
~\\

\textit{Justification}: To evaluate whether the experimental findings are ``somewhat trivial'', I need to assess the novelty, significance, and non-obviousness of the results presented in the paper.\\

**Novel contributions and non-trivial findings:**\\
\hspace{6.2cm} $\dots$

**Overall assessment:**\\

While some high-level expectations (like catastrophic forgetting occurring) are predictable, the specific findings about which methods work best, the quantitative trade-offs, the multi-phase degradation effects, and the detailed analysis of what causes forgetting provide substantial non-trivial insights. The systematic benchmarking and the discovery that traditional CL methods don't transfer well to this domain are significant contributions.\\
~\\

The experimental work is comprehensive, the findings have practical implications for maintaining up-to-date language models, and several results challenge conventional wisdom from traditional continual learning.

\vspace{5pt}
\end{tcblisting}

\vspace{-2mm}
\caption{Example \#1 of ClaimScore evaluation.}
\label{fig:error_analysis_claim_1}
\end{figure*}

\begin{figure*}[!h]
\footnotesize
\begin{tcblisting}{text only,
    halign=left, 
    title=\textbf{Input: Claim (Review Point) \& Paper}, 
    colbacktitle=gray!30!white, 
    coltitle=black
}
\vspace{5pt}

\textit{Paper}:\\
Temperature Schedules for self-supervised contrastive methods on long-tail data~\citep{g4p2}

\hdashrule{\linewidth}{0.4pt}{.5mm} \vspace{-0pt}
\textit{Claim (Review Point)}:\\
The improvement is quite limited compared with other long-tailed learning methods.

\vspace{5pt}
\end{tcblisting}

\begin{tcblisting}{text only,
    halign=left, 
    title=\textbf{Human-annotated Label (5-point Scale)}, 
    colbacktitle=gray!30!white, 
    coltitle=black
}
\vspace{5pt}
2, 2, 3 (Final: 2)
\vspace{5pt}
\end{tcblisting}

\begin{tcblisting}{text only,
    halign=left, 
    title=\textbf{Output: ClaimScore Evaluation},
    colbacktitle=gray!30!white, 
    coltitle=black
}
\vspace{5pt}

\textit{Model}: Claude Sonnet 3.7\\
~\\

\textit{Score}: \textcolor{black}{\textbf{3}}\\
~\\

\textit{Justification}: The proposition states that ``The improvement is quite limited compared with other long-tailed learning methods''. To evaluate this claim, I need to examine how the paper compares its proposed temperature scheduling (TS) approach with other long-tailed learning methods.
Looking at the paper:\\
\hspace{6.2cm} $\dots$

Given that the paper shows consistent improvements over the baselines and SDCLR, but doesn't compare with a broader range of long-tailed learning methods, and the improvements are modest (1-3\%), the proposition that ``the improvement is quite limited compared with other long-tailed learning methods'' is partially supported but not fully substantiated.

\hdashrule{\linewidth}{0.4pt}{.5mm} \vspace{-0pt}
\textit{Model}: Claude Sonnet 4\\
~\\

\textit{Score}: \textcolor{red}{\textbf{4}}\\
~\\

\textit{Human-model Disagreement Type}: \textcolor{red}{Value judgment}\\
~\\

\textit{Justification}: To evaluate this proposition, I need to examine the experimental results and compare the improvements achieved by the proposed temperature scheduling (TS) method against other long-tailed learning methods. Looking at the experimental results:\\
~\\

1. **Comparison with SDCLR**: In Table 3, the paper compares SimCLR+TS with SDCLR (a specialized method for long-tailed contrastive learning). The improvements are modest.\\

2. **Absolute improvements**: The improvements over baseline methods are generally in the range of 1--4\% across different metrics and datasets.\\

3. **Limited baseline comparisons**: The paper primarily compares against standard contrastive learning methods (MoCo, SimCLR) and only one specialized long-tailed method (SDCLR). It doesn't compare against other established long-tailed learning approaches from supervised learning or other self-supervised long-tailed methods.\\

4. **Scope of evaluation**: The evaluation is limited to relatively small datasets (CIFAR variants and ImageNet-100) and doesn't demonstrate performance on larger-scale long-tailed datasets that are commonly used in long-tailed learning literature.\\

5. **Method simplicity vs. improvement**: While the method is simple (just scheduling temperature), the improvements are incremental rather than substantial breakthroughs.\\
~\\

The paper acknowledges this limitation implicitly by positioning the method as ``simple yet effective'' and emphasizing the zero additional computational cost, rather than claiming state-of-the-art performance improvements.

\vspace{5pt}
\end{tcblisting}

\vspace{-2mm}
\caption{Example \#2 of ClaimScore evaluation.}
\label{fig:error_analysis_claim_2}
\end{figure*}

\begin{figure*}[!h]
\footnotesize
\begin{tcblisting}{text only,
    halign=left, 
    title=\textbf{Input: Premise of Argument (Review Point) \& Paper}, 
    colbacktitle=gray!30!white, 
    coltitle=black
}
\vspace{5pt}

\textit{Paper}:\\
Automatic Chain of Thought Prompting in Large Language Models~\citep{g5p1}

\hdashrule{\linewidth}{0.4pt}{.5mm} \vspace{-0pt}
\textit{Argument (Review Point)}:\\
Firstly, when mentioning that the method automatically finds prompts using chain of thoughts (Auto\_CoT) approach, this does not imply finding the best samples from the dataset and using the same method as before. In other words, the proposed method does not find the best chain of thought prompts as claimed but rather believes that ``let's do this step by step'' is the best zero-shot-CoT prompting strategy and the method merely uses that with different prompt samples. The title and the abstract are highly misleading in that way.

\hdashrule{\linewidth}{0.4pt}{.5mm} \vspace{-0pt}
\textit{Premise}:\\
\text{The method uses ``let's do this step by step'' with different prompt samples.}\\

\vspace{5pt}
\end{tcblisting}

\begin{tcblisting}{text only,
    halign=left, 
    title=\textbf{Human-annotated Label (5-point Scale)}, 
    colbacktitle=gray!30!white, 
    coltitle=black
}
\vspace{5pt}
5, 5, 5 (Final: 5)
\vspace{5pt}
\end{tcblisting}

\begin{tcblisting}{text only,
    halign=left, 
    title=\textbf{Output: ClaimScore Evaluation},
    colbacktitle=gray!30!white, 
    coltitle=black
}
\vspace{5pt}

\textit{Model}: Claude Sonnet 3.7\\
~\\

\textit{Score}: \textcolor{black}{\textbf{5}}\\
~\\

\textit{Justification}: The premise states that the Auto-CoT method uses ``let's do this step by step'' with different prompt samples. I need to check if this is accurate based on the paper.
Looking at the paper, in the ``DEMONSTRATION SAMPLING'' section, the author describes how they construct demonstrations:\\
~\\

``A prompted input is formulated as: [Q: q(i)j. A: [P]], where [P] is a single prompt `Let's think step to step'. This formed input is fed into an LLM using Zero-Shot-CoT (Kojima et al., 2022) to output the reasoning chain consisting of the rationale r(i)j and the extracted answer a(i)j.''

\hspace{6.2cm} $\dots$

So the premise is accurate - the Auto-CoT method does use ``let's think step by step'' (or a very similar variant ``Let's think step to step'') with different prompt samples selected through their clustering and sampling approach.

\hdashrule{\linewidth}{0.4pt}{.5mm} \vspace{-0pt}
\textit{Model}: Claude Sonnet 4\\
~\\

\textit{Score}: \textcolor{red}{\textbf{2}}\\
~\\

\textit{Human-model Disagreement Type}: \textcolor{red}{Misunderstanding / incorrect reasoning}\\
~\\

\textit{Justification}: To judge the factuality of this premise, I need to examine what the Auto-CoT method actually does according to the paper.
From the paper, I can see that:\\
~\\

1. Auto-CoT consists of two main stages: (i) question clustering and (ii) demonstration sampling\\
2. In the demonstration sampling stage, the method uses Zero-Shot-CoT to generate reasoning chains for selected questions\\
3. The paper explicitly states that Zero-Shot-CoT uses the prompt ``Let's think step by step'' (not ``let's do this step by step'')\\

\hspace{6.2cm} $\dots$

So the premise is mostly accurate in describing the approach (using Zero-Shot-CoT with different samples), \textcolor{red}{but contains a factual error in the exact wording of the prompt. The paper consistently uses ``Let's think step by step'' rather than ``let's do this step by step.''}\\

\vspace{5pt}
\end{tcblisting}

\vspace{-2mm}
\caption{Example \#1 of ArgScore evaluation.}
\label{fig:error_analysis_arg_1}
\end{figure*}

\begin{figure*}[!h]
\footnotesize
\begin{tcblisting}{text only,
    halign=left, 
    title=\textbf{Input: Premise of Argument (Review Point) \& Paper}, 
    colbacktitle=gray!30!white, 
    coltitle=black
}
\vspace{5pt}

\textit{Paper}:\\
Automatic Chain of Thought Prompting in Large Language Models~\citep{g5p1}

\hdashrule{\linewidth}{0.4pt}{.5mm} \vspace{-0pt}
\textit{Argument (Review Point)}:\\
Secondly, this is very obvious to anyone that the prompts must be representative of the test samples (as much as possible). This work takes the dataset, clusters them, finds the most representative sample from each cluster, and uses that as prompts. There is nothing extra-ordinary about this approach and is merely a small improvement over the randomly sampling strategy. This can also be seen from a marginal improvement in the results over a random sampling strategy.

\hdashrule{\linewidth}{0.4pt}{.5mm} \vspace{-0pt}
\textit{Premise}:\\
The approach is obvious (i.e., prompts must be representative of test samples).

\vspace{5pt}
\end{tcblisting}

\begin{tcblisting}{text only,
    halign=left, 
    title=\textbf{Human-annotated Label (5-point Scale)}, 
    colbacktitle=gray!30!white, 
    coltitle=black
}
\vspace{5pt}
4, 4, 5 (Final: 4)
\vspace{5pt}
\end{tcblisting}

\begin{tcblisting}{text only,
    halign=left, 
    title=\textbf{Output: ClaimScore Evaluation},
    colbacktitle=gray!30!white, 
    coltitle=black
}
\vspace{5pt}

\textit{Model}: Claude Sonnet 3.7\\
~\\

\textit{Score}: \textcolor{red}{\textbf{2}}\\
~\\

\textit{Human-model Disagreement Type}: \textcolor{red}{Misunderstanding / incorrect reasoning}\\
~\\

\textit{Justification}: \textcolor{red}{
I'll examine whether the paper itself acknowledges or implies that prompts should be representative of test samples as an obvious approach:
}\\

\hspace{6.2cm} \textcolor{red}{$\dots$}

\textcolor{red}{The paper's main contribution is showing that diversity in demonstrations is important, not just representativeness. The paper states: ``diversity-based clustering may mitigate misleading by similarity'' and ``diversity of demonstration questions is the key.''}\\
~\\

\textcolor{red}{This suggests that the premise oversimplifies the paper's findings. The paper doesn't claim that prompts must simply be representative -- it argues for a balance of representativeness and diversity.}\\

\hdashrule{\linewidth}{0.4pt}{.5mm} \vspace{-0pt}
\textit{Model}: Claude Sonnet 4\\
~\\

\textit{Score}: \textcolor{black}{\textbf{4}}\\
~\\

\textit{Justification}: To judge the factuality, I need to determine whether it's true that ``prompts must be representative of test samples'' is an obvious principle. This is a general statement about machine learning methodology and prompt engineering principles that would be known to AI practitioners before 2023. The idea that training/demonstration examples should be representative of test data is a fundamental principle in machine learning that predates this specific paper. This principle is well-established in the field and doesn't require knowledge of this specific paper to evaluate its truth.\\
~\\

The paper does implement this principle through clustering and selecting representative samples, but the general principle itself is established knowledge in the field.

\vspace{5pt}
\end{tcblisting}

\vspace{-2mm}
\caption{Example \#2 of ArgScore evaluation.}
\label{fig:error_analysis_arg_2}
\end{figure*}

\begin{figure*}[!h]
\footnotesize
\begin{tcblisting}{text only,
    halign=left, 
    title=\textbf{Input: Question (Review Point) \& Paper}, 
    colbacktitle=gray!30!white, 
    coltitle=black
}
\vspace{5pt}

\textit{Paper}:\\
ReAct: Synergizing Reasoning and Acting in Language Models~\citep{g5p2}

\hdashrule{\linewidth}{0.4pt}{.5mm} \vspace{-0pt}
\textit{Question (Review Point)}:\\
Is there a way to add self-consistency to ReAct, such that it samples multiple reasoning traces and actions and uses the majority action, for example? SC seems extremely effective for Chain of Thought tasks, and it seems like something similar could work here.

\vspace{5pt}
\end{tcblisting}

\begin{tcblisting}{text only,
    halign=left, 
    title=\textbf{Human-annotated Label (5-point Scale)}, 
    colbacktitle=gray!30!white, 
    coltitle=black
}
\vspace{5pt}
2, 2, 5 (Final: 2)
\vspace{5pt}
\end{tcblisting}

\begin{tcblisting}{text only,
    halign=left, 
    title=\textbf{Output: ClaimScore Evaluation},
    colbacktitle=gray!30!white, 
    coltitle=black
}
\vspace{5pt}

\textit{Model}: Claude Sonnet 3.7\\
~\\

\textit{Score}: \textcolor{red}{\textbf{4}}\\
~\\

\textit{Human-model Disagreement Type}: \textcolor{red}{Minority score prediction}\\
~\\

\textit{Justification}: The question asks about adding self-consistency to ReAct, specifically whether it's possible to sample multiple reasoning traces and actions and use the majority action, similar to how self-consistency works for Chain of Thought tasks.
Let me check if the paper discusses this concept or something similar:\\

\hspace{6.2cm} $\dots$

However, the paper does not explicitly discuss applying the self-consistency technique directly to ReAct itself (i.e., sampling multiple ReAct trajectories and taking the majority action). It only discusses combining ReAct with CoT-SC as separate methods, not applying the self-consistency sampling technique to ReAct's own trajectories.

\hdashrule{\linewidth}{0.4pt}{.5mm} \vspace{-0pt}
\textit{Model}: Claude Sonnet 4\\
~\\

\textit{Score}: \textcolor{red}{\textbf{4}}\\
~\\

\textit{Human-model Disagreement Type}: \textcolor{red}{Minority score prediction}\\
~\\

\textit{Justification}: The question asks about adding self-consistency to ReAct, similar to how it's used in Chain of Thought (CoT) tasks. Let me examine what the paper says about self-consistency and whether it discusses or implements this for ReAct.
Looking through the paper, I can see that:\\

\hspace{6.2cm} $\dots$

However, the question specifically asks about applying self-consistency directly to ReAct itself -- sampling multiple ReAct reasoning traces and actions and using majority voting. While the paper discusses CoT-SC and hybrid approaches that combine ReAct with CoT-SC, it doesn't appear to directly implement or discuss a ``ReAct-SC'' approach where multiple ReAct trajectories are sampled and majority voting is applied to the ReAct outputs themselves.

\vspace{5pt}
\end{tcblisting}

\vspace{-2mm}
\caption{Example \#1 of QScore evaluation.}
\label{fig:error_analysis_q_1}
\end{figure*}

\begin{figure*}[!h]
\footnotesize
\begin{tcblisting}{text only,
    halign=left, 
    title=\textbf{Input: Question (Review Point) \& Paper}, 
    colbacktitle=gray!30!white, 
    coltitle=black
}
\vspace{5pt}

\textit{Paper}:\\
Large Language Models Can Self-improve~\citep{g5p8}

\hdashrule{\linewidth}{0.4pt}{.5mm} \vspace{-0pt}
\textit{Question (Review Point)}:\\
Can you check if baseline self-consistency with more examples can match LMSI or not?

\vspace{5pt}
\end{tcblisting}

\begin{tcblisting}{text only,
    halign=left, 
    title=\textbf{Human-annotated Label (5-point Scale)}, 
    colbacktitle=gray!30!white, 
    coltitle=black
}
\vspace{5pt}
4, 5, 5 (Final: 5)
\vspace{5pt}
\end{tcblisting}

\begin{tcblisting}{text only,
    halign=left, 
    title=\textbf{Output: ClaimScore Evaluation},
    colbacktitle=gray!30!white, 
    coltitle=black
}
\vspace{5pt}

\textit{Model}: Claude Sonnet 3.7\\
~\\

\textit{Score}: \textcolor{red}{\textbf{2}}\\
~\\

\textit{Human-model Disagreement Type}: \textcolor{red}{Misunderstanding / incorrect reasoning} \\
~\\

\textit{Justification}: The paper includes a hyperparameter study in Figure 4(b) that shows the effect of varying the number of sampled reasoning paths for self-consistency, both with and without LMSI. \textcolor{red}{This directly addresses whether increasing the number of examples in baseline self-consistency can match LMSI performance.}

\hdashrule{\linewidth}{0.4pt}{.5mm} \vspace{-0pt}
\textit{Model}: Claude Sonnet 4\\
~\\

\textit{Score}: \textcolor{black}{\textbf{4}}\\
~\\

\textit{Justification}: Looking at the experimental setup and results:\\
~- The paper uses m=32 reasoning paths for generating training data\\
~- Figure 4(b) shows accuracy vs number of sampled reasoning paths for both w/o LMSI and LMSI\\
~- The experiments seem to use consistent numbers of paths for comparison\\
~\\

The paper doesn't systematically explore whether dramatically increasing the number of paths in baseline self-consistency (e.g., to 100+ paths) could match the LMSI performance. While Figure 4(b) shows diminishing returns after 15--20 paths, it doesn't explore very high numbers of paths that might be computationally equivalent to the LMSI training process.

\vspace{5pt}
\end{tcblisting}

\vspace{-2mm}
\caption{Example \#2 of QScore evaluation.}
\label{fig:error_analysis_q_2}
\end{figure*}

\clearpage
\section{Model Details}
\label{sec:model-details}
We measure reliability of automatic \textsc{ReviewScore} evaluation with eight current state-of-the-art LLMs, including five proprietary models and three open-sourced models.
For proprietary models, we use
Anthropic's \texttt{claude-3-7-sonnet-20250219}~\citep{claude-3-7-sonnet} and \texttt{claude-sonnet-4-20250514}~\citep{claude-sonnet-4},
OpenAI's \texttt{gpt-4o-2024-08-06}~\citep{gpt4o} and \texttt{gpt-5-2025-08-07}~\citep{gpt5},
and Google's Gemini 2.5 Flash~\citep{gemini}.
Since the \textsc{ReviewScore} evaluation does not require longer chain of thought, we exclude reasoning models. Following this rule, we use Anthropic's claude models without thinking modes and OpenAI's gpt-5 model with minimal reasoning effort.
For open-sourced models, we use
Alibaba's \texttt{qwen3-235b-a22b-2507}~\citep{qwen3} with 235B model parameters,
DeepSeek's \texttt{deepseek-v3-0324}~\citep{deepseek-v3} with 671B model parameters
and Meta's \texttt{llama-3.3-70b-instruct}~\citep{llama3} with 70B model parameters.
We call the models by their APIs.

\vfill

\section{Prompts}
\label{sec:prompts}
We list every prompt used for automatic \textsc{ReviewScore} evaluation (Section \ref{sec:experiments}), automatic review point type classification (used in preprocessing stage of the dataset construction in Section \ref{subsec:dataset}), and automatic argument reconstruction (Section \ref{subsec:auto-argrecon}).
Figure \ref{fig:rubric_unans} -- Figure \ref{fig:prompt_eval_arg_fact} indicate prompts for automatic \textsc{ReviewScore} evaluation and related score rubrics.
Figure \ref{fig:prompt_recon_is_arg} and Figure \ref{fig:prompt_recon_is_q} indicate prompts for automatic review point type classification, where the type is one of claim, argument, and question.
Figure \ref{fig:prompt_recon_verbatim} -- Figure \ref{fig:prompt_recon_faithfulness} indicate prompts for automatic argument reconstruction.
In the prompts, capital letters with double square brackets (i.e., [[XYZ]]) would be replaced by the corresponding material, and words with a curly bracket (i.e., \{abc\}) would be replaced by the corresponding score rubric (Figure \ref{fig:rubric_unans} -- Figure \ref{fig:rubric_untriv}) or a python code snippet in Figure \ref{fig:prompt_recon_validity_code}.

\begin{figure}[!h]
\footnotesize

\begin{tcblisting}{text only,
    halign=left, 
    title=\textbf{Unanswerability Score Rubric}, 
    colbacktitle=gray!30!white, 
    coltitle=black,
}
1: Definitely Answerable by the paper\\
2: Probably Answerable by the paper\\
3: No Verdict\\
4: Probably Unanswerable by the paper\\
5: Definitely Unanswerable by the paper\\
\end{tcblisting}
\vspace{-2mm}
\caption{Score rubric for evaluating \textit{unswerability}.}
\label{fig:rubric_unans}
%
\end{figure}

\begin{figure}[!h]
\footnotesize

\begin{tcblisting}{text only,
    halign=left, 
    title=\textbf{Factuality Score Rubric}, 
    colbacktitle=gray!30!white, 
    coltitle=black,
}
 1: Definitely False\\
 2: Probably False\\
 3: No Verdict\\
 4: Probably True\\
 5: Definitely True
\end{tcblisting}
\vspace{-2mm}
\caption{Score rubric for evaluating \textit{factuality}.}
\label{fig:rubric_fact}
%
\end{figure}

\begin{figure}[!h]
\footnotesize

\begin{tcblisting}{text only,
    halign=left, 
    title=\textbf{Untrivialness Score Rubric}, 
    colbacktitle=gray!30!white, 
    coltitle=black,
}
 0: Definitely Trivial\\
 1: Probably Trivial\\
 2: Definitely Not Trivial
\end{tcblisting}
\vspace{-2mm}
\caption{Score rubric for evaluating \textit{untrivialness}.}
\label{fig:rubric_untriv}
%
\end{figure}

\begin{figure*}[!h]
\footnotesize

\begin{tcblisting}{text only,
    halign=left, 
    title=\textbf{Prompt for evaluating QScore (\textit{unanswerability of questions})}, 
    colbacktitle=gray!30!white, 
    coltitle=black,
}

\# Paper\\
{[[PAPER]]}\\
~\\

\# Question\\
{[[QUESTION]]}\\
~\\

You are given a paper submitted to an AI conference and a question regarding the paper.\\
~\\

Judge if the question is answerable or not by the paper. You should scale 1-5 to indicate unanswerability as follows.\\
\{Unanswerability Score Rubric\}\\
~\\

If you score the question 1 or 2, then indicate which knowledge source you have grounded to (i.e., indicate corresponding section(s) and verbatim sentence(s)) and answer the question.\\
~\\

Your output should be the following.\\
~\\

\#\#\# Reasoning\\
{[think step-by-step]}\\
~\\

\#\#\# Unanswerability\\
{[1 or 2 or 3 or 4 or 5]}\\
~\\

\#\#\# Source\\
{[corresponding section(s) and verbatim sentence(s) if the score is 1 or 2, otherwise None]}\\
~\\

\#\#\# Answer\\
{[answer the question if the score is 1 or 2, otherwise None]}
\end{tcblisting}
\vspace{-2mm}
\caption{Prompt used for evaluating QScore in a 5-point scale.}
\label{fig:prompt_eval_q_unans}
%
\end{figure*}

\begin{figure*}[!h]
\footnotesize

\begin{tcblisting}{text only,
    halign=left, 
    title=\textbf{Prompt for evaluating ClaimScore or WScore (\textit{factuality of weaknesses})}, 
    colbacktitle=gray!30!white, 
    coltitle=black,
}

\# Paper\\
{[[PAPER]]}\\
~\\

\# Weakness\\
{[[WEAKNESS]]}\\
~\\

You are given a paper submitted to an AI conference and a weakness regarding the paper.\\
~\\

Judge if the weakness is true or not based on the paper. You should scale 1-5 to indicate factuality as follows.\\
\{Factuality Score Rubric\}\\
~\\

Your output should be the following.\\
~\\

\#\#\# Reasoning\\
{[think step-by-step]}\\
~\\

\#\#\# Factuality\\
{[1 or 2 or 3 or 4 or 5]}\\
\end{tcblisting}
\vspace{-2mm}
\caption{Prompt used for evaluating ClaimScore or WScore in a 5-point scale.}
\label{fig:prompt_eval_w_fact}
\vspace{-4mm}
\end{figure*}

\begin{figure*}[!h]
\footnotesize
\vspace{-13mm}
\begin{tcblisting}{text only,
    halign=left, 
    title=\textbf{Prompt for evaluating ArgScore (\textit{factuality of premises})}, 
    colbacktitle=gray!30!white, 
    coltitle=black,
}

\# Paper\\
{[[PAPER]]}\\
~\\

\# Weakness\\
{[[WEAKNESS]]}\\
~\\

\# Premise\\
{[[PREMISE]]}\\
~\\

You are given a paper submitted to an AI conference, a weakness of the paper, and one of premises of the weakness. Your task is to judge the factuality and untrivialness of the given premise.\\
~\\

First, judge the factuality of the premise. To do that, choose an appropriate knowledge source from:\\
~~~1. given paper\\
~~~2. annotator knowledge before the year [[YEAR]] (more precisely, before the paper is publicized)\\
~~~3. other paper(s),\\
and then judge the factuality of the premise based on the knowledge source. You should scale 1-5 to indicate factuality as follows.\\
\{Factuality Score Rubric\}\\
~\\

Here are guidelines you should follow:\\
~- Main purpose is to distinguish \text{*given\_paper*} and \text{*annotator\_knowledge*}.\\
~- Select \text{*other\_papers*} only if the premise refers a specific paper.\\
~- Note that you should separate judging the factuality of the premise from understanding the semantics of the premise. It does not matter whether \text{*given\_paper*} is needed or not to understand the semantics of the premise. The knowledge source is \text{*given\_paper*} only if \text{*given\_paper*} is needed to judge the factuality of the premise, otherwise, the knowledge source is \text{*annotator\_knowledge*}. For example, although you need the paper's context in order to understand what the premise means, if you do not need the paper's knowledge to judge if the premise is true or not (e.g., logical assessment), then you should choose \text{*annotator\_knowledge*} as a knowledge source and judge the factuality accordingly.\\
~- For premises that are conditionals (If A then B), you should presume that the antecedent (A) is always true even if the antecedent does not align with the paper's knowledge. (Because the antecedent is always true, the knowledge source should only be determined while judging the factuality of the conseqeunt.) Then, choose an appropriate knowledge source to judge if the consequent (B) is true or not and judge the factuality accordingly.\\
~\\

Next, decide whether the premise is trivially true or not based on the common knowledge of CS/AI-majoring undergrad students before the year {[[YEAR]]} (more precisely, before the paper is publicized). For premises that are conditionals (If A then B), you must assume that the antecedent (A) is true and judge if the consequent (B) is trivially true or not. You should scale the score to 0-2 as follows.\\
\{Untrivialness Score Rubric\}\\
~\\

Here are guidelines you should follow:\\
 - If the knowledge source is \text{*given\_paper*} or \text{*other\_papers*}, then untrivialness should always be 2 unless the premise factuality could also be determined by \text{*annotator\_knowledge*}.\\
 - If the knowledge source is \text{*annotator\_knowledge*}, then untrivialness could be 0-2.\\
~\\

Your output should be formatted as below.\\
~\\

\#\#\# Reasoning\\
{[think step-by-step]}\\
~\\

\#\#\# Source\\
{[given\_paper or annotator\_knowledge or other\_papers]}\\
~\\

\#\#\# Factuality\\
{[1 or 2 or 3 or 4 or 5]}\\
~\\

\#\#\# Untrivialness\\
{[0 or 1 or 2]}
\end{tcblisting}
\vspace{-2mm}
\caption{Prompt used for evaluating ArgScore in a 5-point scale.}
\label{fig:prompt_eval_arg_fact}
\vspace{-4mm}
\end{figure*}

\begin{figure*}[!h]
\footnotesize

\begin{tcblisting}{text only,
    halign=left, 
    title=\textbf{Prompt for classifying review point types (\texttt{is\_argument})}, 
    colbacktitle=gray!30!white, 
    coltitle=black,
}
\# Review Point\\
{[[REVIEW\_POINT]]}\\
~\\

You are given an AI conference review point. Is this an argument or not? Your response should follow the format below.\\
~\\

\#\#\# Reasoning\\
{[think step-by-step]}\\
~\\

\#\#\# Response\\
{[Yes or No]}
\end{tcblisting}
\vspace{-2mm}
\caption{Prompt used for deciding if a review point is an argument or not (\texttt{is\_argument}).}
\label{fig:prompt_recon_is_arg}
%
\end{figure*}

\begin{figure*}[!h]
\footnotesize

\begin{tcblisting}{text only,
    halign=left, 
    title=\textbf{Prompt for classifying review point types (\texttt{is\_question})}, 
    colbacktitle=gray!30!white, 
    coltitle=black,
}
\# Review Point\\
{[[REVIEW\_POINT]]}\\
~\\

You are given an AI conference review point. Decide if this is a question or a simple claim. Your response should follow the format below.\\
~\\

\#\#\# Reasoning\\
{[think step-by-step]}\\
~\\

\#\#\# Response\\
{[Question or Claim]}
\end{tcblisting}
\vspace{-2mm}
\caption{Prompt used for deciding if a review point is a question or a claim (\texttt{is\_question}).}
\label{fig:prompt_recon_is_q}
%
\end{figure*}

\begin{figure*}[!h]
\footnotesize

\begin{tcblisting}{text only,
    halign=left, 
    title=\textbf{Prompt for Argument Reconstruction (\texttt{extract\_verbatim\_conclusion\_reason})}, 
    colbacktitle=gray!30!white, 
    coltitle=black,
}
\#\#\# Review Point\\
{[[REVIEW\_POINT]]}\\
~\\

Given an AI conference review point, consider it as an argument, and then returns its verbatim conjecture in the source text and verbatim reason statements of that conjecture in the source text. The output format should be as following.\\
~\\

\#\#\# Conjecture\\
{[main conjecture in the review point]}\\
~\\

\#\#\# Supporting Reasons\\
{[list of supporting reasons for the conjecture]}
\end{tcblisting}
\vspace{-2mm}
\caption{Prompt used for extracting verbatim conjecture and reason statements in an argument (\texttt{extract\_verbatim\_conclusion\_reason}).}
\label{fig:prompt_recon_verbatim}
%
\end{figure*}

\begin{figure*}[!h]
\footnotesize

\begin{tcblisting}{text only,
    halign=left, 
    title=\textbf{Prompt for Argument Reconstruction (\texttt{argument\_reconstruction})}, 
    colbacktitle=gray!30!white, 
    coltitle=black,
}
\# Paper\\
{[[PAPER]]}\\
~\\

\# Review Point\\
~\\

\#\# Conclusion\\
{[[CONCLUSION]]}\\
~\\

\#\# Explicit reasons\\
{[[REASONS]]}\\
~\\

You are given a paper submitted to an AI conference and a review point by a peer reviewer. A review point consists of a conclusion and its explicit reasons. Reconstruct an argument (i.e., a review) with premise-conclusion structure where premises deductively imply the conclusion. The reconstructed argument should be deductively valid, using formal logical patterns like modus ponens (e.g., Premise1: A, Premise2: If A then B, Conclusion: B). Add implicit premises and intermediate conclusions if needed.\\
~\\

Your output should composed of two parts, argument reconstruction and its formalization. In the first part, list premises, intermediate conclusions, and conclusion, and indicate their logical connection (i.e., which propositions deductively implies which). In the second part, first define variables and/or predicates, then formalize premises, intermediate conclusions, and a conclusion, and then generate a deductive proof. The output format should be as following.\\
~\\

\# Argument Reconstruction\\
~\\

\#\# Premises\\
{[list of explicit and implicit premises]}\\
~\\

\#\# Intermediate Conclusions\\
{[list of intermediate conclusions (if intermediate conclusions are not needed, then write ``None''.)]}\\
~\\

\#\# Conclusion\\
{[a conclusion]}\\
~\\

\#\# Logical Connections\\
{[list of logical connections]}\\
~\\

\# Formalized Argument\\
~\\

\#\# Defined Variables/Predicates\\
{[definition of each variable and/or predicate]}\\
~\\

\#\# Formalized Premises\\
{[formalization of premises using definition]}\\
~\\

\#\# Formalized Intermediate Conclusions\\
{[formalization of intermediate conclusions using definition (if intermediate conclusions are not needed, then write ``None''.)]}\\
~\\

\#\# Formalized Conclusion\\
{[formalization of conclusion using definition]}\\
~\\

\#\# Deductive Proof\\
{[deductive proof using formalized premises]}
\end{tcblisting}
\vspace{-2mm}
\caption{Prompt used for reconstructing an argument (\texttt{argument\_reconstruction}).}
\label{fig:prompt_recon_recon}
%
\end{figure*}

\begin{figure*}[!h]
\footnotesize

\begin{tcblisting}{text only,
    halign=left, 
    title=\textbf{Prompt for Argument Reconstruction (\texttt{streamlining})}, 
    colbacktitle=gray!30!white, 
    coltitle=black,
}
\#\# Defined Variables/Predicates\\
{[[DEFINITION]]}\\
~\\

\#\# Formalized Premises\\
{[[PREMISES]]}\\
~\\

\#\# Formalized Conclusion\\
{[[CONCLUSION]]}\\
~\\

\#\# Deductive Proof\\
{[[PROOF]]}\\
~\\

First, determine necessary formalized premises for the given deductive proof. This includes:\\
~1. Add any missing formalized premises that are necessary to prove conclusion but cannot be dervied from the formalized premises.\\
~2. Remove any unnecessary formalized premises that are not necessary to prove conclusion but present in the formalized premises.\\
You should format these premises into a python dictionary where keys and values are python strings.\\
~\\

Second, write a python program using z3 that inputs the necessary formalized premises and formalized conclusion and outputs:\\
~1. Their validity, formatted as a python string of either ``valid'' or ``invalid''.\\
~2. A smallest subset of necessary formalized premises to prove the formalized conclusion, formatted as a python list of keys of the python dictionary of the necessary formalized premises.\\
You should therefore print two things (a python string and a python list) separately. Please use the below python code snippet.\\
\{Code snippet for checking validity\}\\
~\\

Third, return the final formalized conclusion that is used in the python program in step 2.\\
~\\

Lastly, judge whether the formal proof using the necessary formalized premises (in step 1) and the final formalized conclusion (in step 3) is circular or not. If there is a single necessary formalized premise that is the same as the final formalized conclusion, then return N/A.\\
~\\

Your response should be as following.\\
~\\

\#\#\# Necessary Formalized Premises\\
\verb|```|~python\\
\{\\
~~~~``[Symbol of a premise \#1]'': ``{[Formalization of a premise \#1]}'',\\
~~~~``[Symbol of a premise \#2]'': ``{[Formalization of a premise \#2]}'',\\
~~~~...\\
\}\\
\verb|```|\\
~\\

\#\#\# Python Program\\
\verb|```|~python\\
{[a python program]}\\
\verb|```|~\\
~\\

\#\#\# Final Formalized Conclusion\\
{[Formalized conclusion in the python program]}\\
~\\

\#\#\# Proof Circularity\\
{[Yes or No or N/A]}
\end{tcblisting}
\vspace{-2mm}
\caption{Prompt used for streamlining formalized reconstruction (\texttt{streamlining}).}
\label{fig:prompt_recon_validity}
%
\end{figure*}

\begin{figure*}[!h]
\footnotesize

\begin{tcblisting}{text only,
    halign=left, 
    title=\textbf{Code snippet for checking validity}, 
    colbacktitle=gray!30!white, 
    coltitle=black,
}
\verb|```|~python\\
from z3 import *\\
import itertools\\
~\\

\#\#\#\#\#\#\#\#\#\#\#\#\#\#\#\#\#\#\#\#\#\#\#\#\#\#\#\#\#\#\#\#\#\\
\#\#\# Write down your code here \#\#\#\\
\#\#\#\#\#\#\#\#\#\#\#\#\#\#\#\#\#\#\#\#\#\#\#\#\#\#\#\#\#\#\#\#\#\\
~\\

\# Check validity of the argument\\
def check\_validity(premises\_dict, conclusion):\\
~~~~s = Solver()\\
~~~~s.add(list(premises\_dict.values()))\\
~~~~s.add(Not(conclusion))\\
~~~~if s.check() == unsat:\\
~~~~~~~~return ``valid''\\
~~~~else:\\
~~~~~~~~return ``invalid''\\
~\\

\# Find minimal set of premises\\
def find\_minimal\_premises(premises\_dict, conclusion):\\
~~~~for subset\_size in range(1, len(premises\_dict) + 1):\\
~~~~~~~~for subset in itertools.combinations(premises\_dict.keys(), subset\_size):\\
~~~~~~~~~~~~subset\_premises = [premises\_dict[key] for key in subset]\\
~~~~~~~~~~~~if check\_validity(subset\_premises, conclusion) == ``valid'':\\
~~~~~~~~~~~~~~~~return list(subset)\\
~~~~return list(premises\_dict.keys())\\
~\\

validity = check\_validity(premises, conclusion)\\
print(validity)\\
minimal\_premises = find\_minimal\_premises(premises, conclusion)\\
print(minimal\_premises)\\
\verb|```|~\\
~\\
\end{tcblisting}
\vspace{-2mm}
\caption{Python code snippet for evaluating validity of reconstruction.}
\label{fig:prompt_recon_validity_code}
%
\end{figure*}

\begin{figure*}[!h]
\footnotesize

\begin{tcblisting}{text only,
    halign=left, 
    title=\textbf{Prompt for Argument Reconstruction (\texttt{program\_debugging})}, 
    colbacktitle=gray!30!white, 
    coltitle=black,
}
When I execute the python program, I got the following error:\\
{[[ERROR]]}\\
~\\

Fix the error and generate a revised python program. Your response should be as following.\\
~\\

\#\#\# Reasoning\\
{[explain why and how to fix the program]}\\
~\\

\#\#\# Revised Python Program\\
\verb|```|~python\\
{[a python program]}\\
\verb|```|
\end{tcblisting}
\vspace{-2mm}
\caption{Prompt used for debugging python programs that evaluate validity of reconstruction (\texttt{program\_debugging}).}
\label{fig:prompt_recon_debug}
%
\end{figure*}

\begin{figure*}[!h]
\footnotesize

\begin{tcblisting}{text only,
    halign=left, 
    title=\textbf{Prompt for Argument Reconstruction (\texttt{deformalization})}, 
    colbacktitle=gray!30!white, 
    coltitle=black,
}
\#\# Defined Variables/Predicates\\
{[[DEFINITION]]}\\
~\\

\#\# Formalized Premises\\
{[[PREMISES]]}\\
~\\

\#\# Formalized Conclusion\\
{[[CONCLUSION]]}\\
~\\

Given definitions of variables and/or predicates, generate natural language (NL) descriptions of formalized premises and conclusion. Your response should be as following.\\
~\\

\#\#\# NL Premises\\
{[list of premises in natural language]}\\
~\\

\#\#\# NL Conclusion\\
{[conclusion in natural language]}
\end{tcblisting}
\vspace{-2mm}
\caption{Prompt used for translating FOL formulas with keys (i.e., defined variables/predicates) to NL reconstructed arguments (\texttt{deformalization}).}
\label{fig:prompt_recon_deformalization}
%
\end{figure*}

\begin{figure*}[!h]
\footnotesize

\begin{tcblisting}{text only,
    halign=left, 
    title=\textbf{Prompt for Argument Reconstruction (\texttt{check\_faithfulness})}, 
    colbacktitle=gray!30!white, 
    coltitle=black,
}
\# Argument\\
{[[ARGUMENT]]}\\
~\\

\# Argument Reconstruction\\
~\\

\#\# Premises\\
{[[PREMISES]]}\\
~\\

\#\# Conclusion\\
{[[CONCLUSION]]}\\
~\\

For an argument, its reconstruction as a premise-conclusion structure is given. Your task is to judge whether the construction is faithful or not. You should judge the faithfulness according to the following two criteria:\\
~- **Accuracy \& Charity.** The reconstruction should keep the author’s intended meaning while eliminating irrelevancies—i.e., obey the principle of charity and prefer the strongest sensible reading of ambiguous passages.\\
~- **Completeness.** All explicit premises, the main conclusion and any indispensable implicit premises must be included.\\
~\\

The output format should be as following.\\
~\\

\# Reasoning\\
{[Explain step-by-step]}\\
~\\

\# Faithfulness\\
{[Yes or No]}
\end{tcblisting}
\vspace{-2mm}
\caption{Prompt used for evaluating faithfulness of reconstruction (\texttt{check\_faithfulness}).}
\label{fig:prompt_recon_faithfulness}
%
\end{figure*}

\end{document}